\documentclass[journal]{IEEEtran}

\ifCLASSINFOpdf
\else
\fi

\usepackage{mathtools}
\usepackage[english]{babel}
\usepackage{graphicx}
\usepackage{subfigure}
\usepackage{amsmath,amsfonts,amssymb}
\usepackage{algorithm}
\usepackage{algpseudocode}
\usepackage{epstopdf}
\usepackage{color}
\usepackage[normalem]{ulem}
\usepackage{amsthm}
\usepackage{amsbsy}
\usepackage{varioref}
\usepackage{cases}
\usepackage{verbatim}
\usepackage{cite}
\usepackage{breqn}
\usepackage{dblfloatfix}
\usepackage{multirow}
\usepackage{booktabs}

\begin{document}

\title{  Multi-Modal Dictionary Learning for  Image Separation With Application In Art Investigation}


\author{Nikos~Deligiannis,~\IEEEmembership{Member,~IEEE,}
        Jo\~ao~F.~C.~Mota,~\IEEEmembership{Member,~IEEE,}
        Bruno~Cornelis,~\IEEEmembership{Member,~IEEE,}
        Miguel~R.~D.~Rodrigues,~\IEEEmembership{Senior Member,~IEEE,}
        and~Ingrid~Daubechies,~\IEEEmembership{Fellow,~IEEE}
\thanks{
The work is supported by the VUB  research programme M3D2, the EPSRC grant EP/K033166/1, and the VUB-UGent-UCL-Duke International Joint Research Group  (grant VUB: DEFIS41010). A preliminary version of this work is accepted for presentation at the IEEE Int. Conf. Image Process. (ICIP)\ 2016 \cite{deligiannis2016xray}.   

N. Deligiannis and B. Cornelis are with the Department of Electronics
and Informatics, Vrije Universiteit Brussel, Brussels 1050, Belgium,  and  with iMinds, 9050 Ghent, Belgium  (e-mail: \textbraceleft ndeligia, bcorneli\textbraceright@etro.vub.ac.be).
B. Cornelis was with the Department of Mathematics, Duke University, Durham, NC 27708 USA.

J. Mota and M. Rodrigues are with the Department of Electronic and Electrical Engineering, University College London, UK (e-mail: {j.mota,m.rodrigues}@ucl.ac.uk).


I. Daubechies is with the Departments of Mathematics and Electronic and Computer Engineering, Duke University, Durham, NC 27708 USA (e-mail: ingrid@math.duke.edu).}}



\maketitle

\begin{abstract}
In support of art investigation, we propose a new source separation  method that unmixes a single X-ray scan acquired from double-sided paintings. In this problem, the X-ray signals to be separated have similar morphological characteristics, which brings previous source separation methods to their limits. Our solution is to use photographs taken from
the front- and  back-side of the panel to drive the separation process.
 The crux of our approach relies on the coupling of the two imaging modalities (photographs and X-rays) using a novel coupled dictionary learning framework able to capture  both common and disparate features across the modalities using parsimonious representations; the common component models features shared by the multi-modal images, whereas the innovation component captures  modality-specific information. As such, our model enables the formulation of appropriately regularized convex optimization procedures that lead to the accurate separation of the X-rays. Our dictionary learning framework can be tailored  both  to a single-  and a multi-scale framework, with the latter  leading to a significant performance improvement. Moreover, to improve further on the visual quality of the separated images, we propose to train coupled dictionaries that ignore certain parts of the painting corresponding to craquelure. Experimentation on synthetic and real data---taken from digital acquisition of the Ghent Altarpiece (1432)---confirms the superiority of our method against the state-of-the-art morphological component analysis technique that uses either fixed or trained dictionaries to perform image separation. 
\end{abstract}



\IEEEpeerreviewmaketitle

\section{Introduction}

\IEEEPARstart{B}{ig data} sets---produced by scientific experiments or projects---often contain  heterogeneous data obtained by capturing a physical process or object using diverse sensing modalities~\cite{chen2014big}. The result is a rich set of  signals, heterogeneous in nature but  strongly correlated due to their being generated by a common underlying phenomenon.
Multi-modal signal  processing and analysis is thus gaining momentum in various research disciplines ranging from medical diagnosis \cite{zhang2012multi} to remote sensing and computer vision \cite{wang2012semi}.
In particular, the analysis of high-resolution multi-modal digital acquisitions of paintings
in support of art scholarship has proved a challenging
field of research. Examples include the numerical characterization of brushstrokes~\cite{johnson2008,Noord2015}
for the authentication or dating of paintings,  canvas thread counting~\cite{JohnsonJohnson:09,Yangetal2015,Maaten2015}
with applications in art forensics, and the (semi-) automatic detection and
digital inpainting of cracks~\cite{cornelis2012crack,cornelis2013bayesian,pizurica2015digital}.

The \textit{Lasting Support} project has focused on the investigation of the \emph{Ghent Altarpiece} (1432), also known as \emph{The Adoration of the Mystic Lamb}, a polyptych  on wood panel painted by Jan and Hubert van Eyck. One of the most admired and influential masterpieces in the history of art, it has given rise to many puzzling questions for art historians. 
Currently, the \emph{Ghent Altarpiece} is undergoing a major conservation and restoration campaign that is planned to end in 2017. The panels of the masterpiece were documented with various imaging modalities, amongst which visual macrophotography, infrared macrophotography,
infrared reflectography and X-radiography \cite{pizurica2015digital}. A massive visual data set (comprising over 2TB of data) has been compiled by capturing small areas of the
polyptych separately and  stitching the resulting image blocks into one image
per panel \cite{cornelis2011digital}.


\begin{figure}[t]
        \centering
\includegraphics[width=0.5\textwidth]{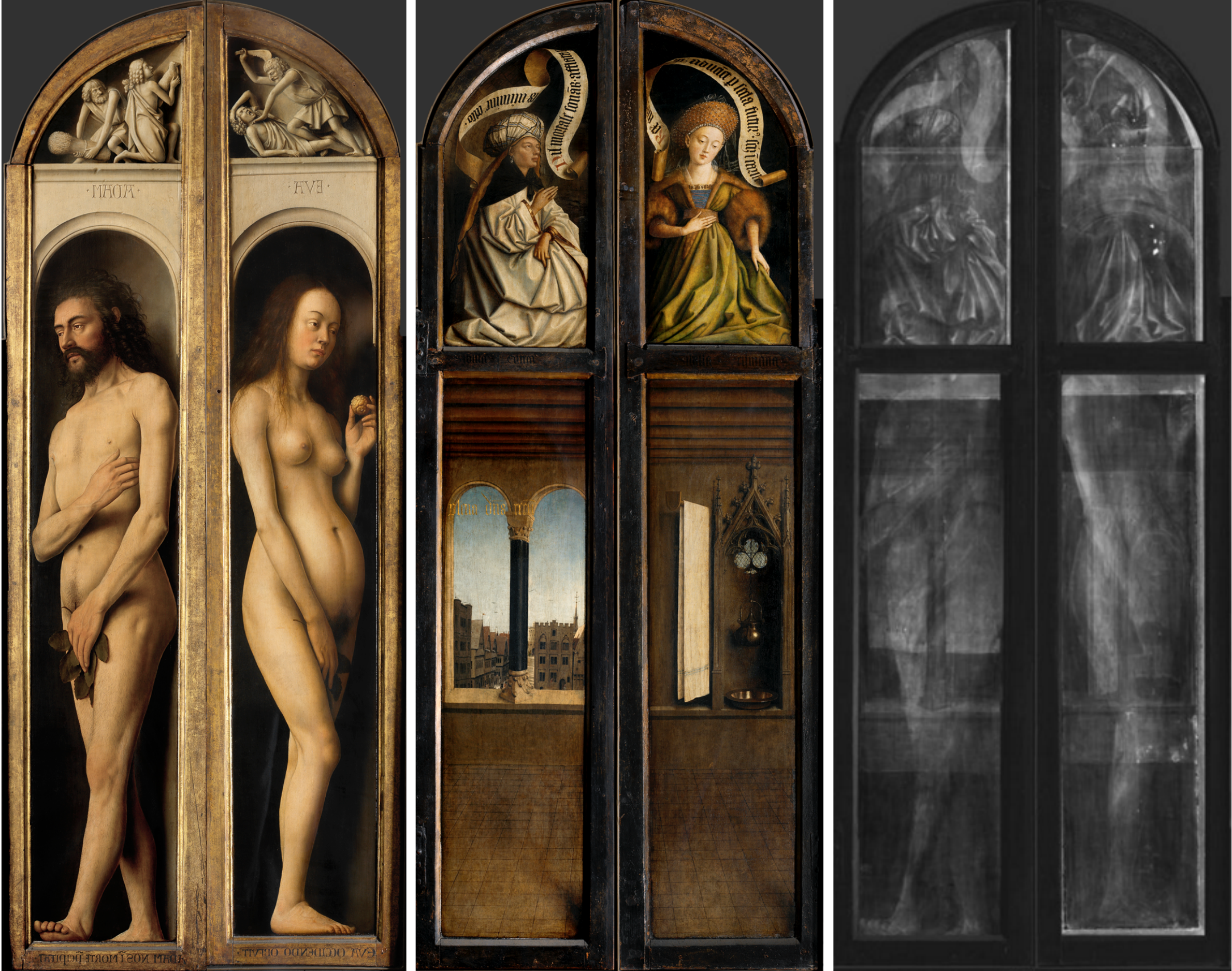}
        \caption{Panels from the \emph{Ghent Altarpiece}:  (left) panels of \emph{Adam} and \emph{Eve}, (centre) the
respective paintings on the back, (right) corresponding X-ray images  containing a mixture of components. \tiny\copyright KIK-IRPA}
        \label{Fig:ghentaltarpiece}
\end{figure}

X-ray images are common tools in  painting investigation, since they reveal information about the composition of the  materials,
the variations in paint thickness,  the support  as well as the cracks and losses in the ground and paint layers.
The problem we address in this paper relates to the outer side panels, namely, the panels showing near life-sized depictions of Adam and Eve, shown in Fig.~\ref{Fig:ghentaltarpiece}. Due to X-ray penetration, the scans of  these panels are a mixture of the paintings from each side of the panel as well as the wood panel itself. The presence of all these components makes the reading of the X-ray image difficult for art experts, who would welcome an effective approach to separate the
components. 



The task of separating a mixture of  signals into its constituent  components
is a popular field of research. Most work addresses the blind source separation
(BSS) problem, where the goal is to retrieve the different sources from a given linear mixture. Several methods attempt to solve the BSS\ problem by imposing  constraints on the sources' structure. Independent component analysis (ICA) \cite{hyvarinen2004independent} commonly assumes that the components are independent non-Gaussian  and attempts to separate them by minimizing the mutual information \cite{hyvarinen1999gaussian}. Nonnegative matrix factorization 
is another approach to solve the problem, where it is assumed that the sources are nonnegative (or they are transformed to a  nonnegative representation)  \cite{smaragdis2014static}.    
In an alternative path, the  problem has been cast into a Bayesian framework, where either the sources are viewed as latent variables  \cite{miskin2000ensemble}, or the  problem is solved by maximizing the joint posterior density of the sources  \cite{rowe2002bayesian}. 
Under the Bayesian methodology, spatial smoothing priors (via, for example, Markov random fields) have been  used to regularize blind image separation problems \cite{kayabol2009bayesian}. These assumptions do not fit our particular problem as both components have similar statistical properties and they are certainly not statistically independent. 
  
Sparsity is another  source prior heavily exploited in BSS \cite{bobin2007sparsity,zibulevsky2001blind}, as well as in various other inverse problems, such as, compressed sensing \cite{candes2006robust,donoho2006compressed}, image  inpainting \cite{guillemot2014image,mairal2008learning}, denoising \cite{elad2006image}, and deconvolution \cite{daubechies2003iterative}. Morphological component analysis (MCA), in particular, is a state-of-the-art sparsity-based regularization method,   initially designed for the single-mixture problem \cite{starck2004redundant,bobin2007sparsity} and then extended to the multi-mixture case \cite{bobin2006morphological}.
The crux of the method is the basic assumption that each component has its own characteristic morphology; namely, each has a highly sparse representation over a   set of bases (or, \textit{dictionaries}), while being highly non-sparse  for the  dictionaries of the other components. 
Prior work in digital painting analysis has employed MCA to remove cradling artifacts within X-ray images of paintings on a panel~\cite{icip14}. The cradling and painting components have very different morphologies, captured by different  predefined dictionaries. Namely, complex wavelets \cite{1998dualtree} provide a sparse representation for the smooth X-ray image and  shearlets \cite{guo2007optimally}  were used to represent the texture of the wood grain. Alternatively,  dictionaries can  be  learned from a set of training signals; several algorithms have been proposed to construct dictionaries including the method of optimal directions (MOD) \cite{engan1999method} and the K-SVD algorithm \cite{aharon2006img}. Both utilize the orthogonal matching pursuit (OMP)~\cite{tropp2007signal} method for sparse decomposition but they differ in the way they update the dictionary elements while learning. Recently, multi-mixture MCA has been combined with K-SVD, resulting in  a method where dictionaries are learned adaptively while separating \cite{abolghasemi2012blind}.

However,  in our particular separation problem we have a simple mixture of two X-ray components that
are morphologically very similar [see Fig. \ref{Fig:ghentaltarpiece}]. Hence, as we will show in the experimental section, simply using fixed or learned dictionaries is insufficient to discriminate one component from the other.  
Unlike prior work, in our setup we have access to high-quality photographic material from each side of the panel that can be used to assist the X-ray image separation process. 

In this work, we elaborate on a novel method to perform  separation of X-ray images from a single mixture by using images of another modality as side information. Our  contributions are as follows: \begin{itemize}
\item We present a new model based on parsimonious representations, which captures both the inherent similarities and the discrepancies among heterogeneous correlated data. The model decomposes the data into a sparse component that is common to the different modalities and a sparse component that uniquely describes each data type. Our model enables the formulation
of appropriately regularized convex optimization procedures that address the separation problem at hand.  
\item We propose a novel dictionary learning approach that trains dictionaries coupling the images from the different modalities. Our approach introduces a new modified OMP\ algorithm that is tailored to our data model.     
\item We devise a novel method that ignores  craquelure pixels---namely, pixels that visualize cracks in the surface of paintings---when learning coupled dictionaries.  Paying no heed to these pixels avoids  contaminating the dictionaries with  high frequency
noise, thereby leading to higher separation performance. Our approach bears similarities with inpainting approaches, e.g., \cite{mairal2008learning}; it is,  however, different in the way the dictionary learning problem is posed and solved. \item We devise a novel multi-scale image separation strategy that is based on a recursive decomposition of the mixed X-ray and
visual images into low-
and high-pass bands. As such, the method enables the accurate separation of high-resolution images even when a local  sparsity prior is assumed. Our approach differs from existing multi-scale dictionary learning methods
\cite{yan2013nonlocal,ophir2011multi,mairal2008learning} not only by considering imaging data gleaned from diverse modalities but also in the way the multi-scale decomposition
is  constructed.
\item We conduct  experiments using synthetic and real  data proving that the use of side information is crucial in the separation of X-ray images from double-sided paintings. 
\end{itemize}

In the remainder of the paper: Section \ref{eq:RelatedWork} reviews related work and Section \ref{sec:SourceSepSideInfor} poses our  source separation with side information problem. Section \ref{sec:coupledDicLearn}  describes the proposed coupled dictionary learning algorithm. Section \ref{sec:weightedcoupledDicLearn} presents our  method that ignores cracks when learning dictionaries and, Section \ref{sec:MultScAppr} elaborates on our single- and multi-scale approaches to X-ray image separation. Section \ref{Sec:ExpSection} presents the  evaluation of  our algorithms while, Section \ref{sec:conclusion} draws our conclusions.  

\section{Related Work}
\label{eq:RelatedWork}
\subsection{Source Separation}
Adhering to a  formal definition, MCA \cite{starck2004redundant,bobin2007sparsity}  decomposes a source
or image mixture $x=\sum_{i=1}^{\kappa} x_{i}$, with $x,x_i\in\mathbb{R}^{n\times
1}$, into its constituents, with the assumption that each $x_{i}$ admits
a sparse decomposition in a different overcomplete dictionary $\Phi_{i}\in\mathbb{R}^{n\times d_{i}}$, $(n\ll\ d_{i})$. Namely, each component can be expressed as $x_i
= \Phi_i z_i$, where $z_i\in\mathbb{R}^{d_{i}\times1}$ is a sparse vector
comprising a few non-zero coefficients: $\|z_i\|_0=\#\{ \xi:z_{i_\xi}\neq0,\xi=1,\dots,d_i\}=s_i\ll
d_i$, with  $\|\cdot\|_0$ denoting the $\ell_0$ pseudo-norm. The BSS problem  is thus addressed as \cite{starck2004redundant,bobin2007sparsity}
\begin{equation}
(\hat{z}_1,\dots,\hat{z}_\kappa) = \arg\min_{z_1,\dots,z_\kappa}\sum_{i=1}^\kappa\|z_i\|_0\;\;\text{s.t.}\;\;
x=\sum_{i=1}^{\kappa} \Phi_i z_i.
\end{equation}

Unlike the BSS problem, informed source separation (ISS) methods utilise some form of prior information to aid the  task at hand. ISS methods are tailored to  the application they address (to the best of our  knowledge they are  applied only for audio mixtures \cite{LiutkusPBGR12,gorlow2013informed}).  For instance, an encoding/decoding framework is proposed in~\cite{LiutkusPBGR12}, where the sources are mixed at the encoder and the mixtures are sent to the decoder together with  side information that is embedded  by means of quantization index modulation (QIM) \cite{Chen99quantizationindex}. Unlike these methods, we propose a generic source separation framework that incorporates side information gleaned from a correlated heterogeneous source by means of a new dictionary learning method that couples the heterogenous sources.

\subsection{Dictionary Learning}
Dictionary learning factorizes a matrix composed of training signals $X=[x_1,\dots,x_k]\in\mathbb{R}^{n\times
k}$ into the product $\Phi Z$ as
\begin{equation}
\label{Eq:DLDefinition}
\Big(\Phi,Z\Big) = \arg\min_{\Phi',Z'} \|X-\Phi' Z'\|_F^2\;\;\text{s.t.}\;
\|z_i\|_0\leq s,\;i=1,\dots,k,
\end{equation}
where $Z=[z_1,\dots,z_k]\in R^{d \times k}$ contains the sparse vectors corresponding
to the signals $X=[x_1,\dots,x_k]$ and $\|\cdot\|_F$ is the Frobenius
norm of a matrix. The columns of the dictionary $\Phi$
are typically constrained to have unit norm so as to improve the identifiability of the dictionary. To solve Problem \eqref{Eq:DLDefinition}, which is non-convex, Olshausen and Field \cite{olshausen1996emergence} proposed to iterate
through a step that learns the sparse codes 
and a step that updates the dictionary elements. The same strategy is  followed in subsequent studies \cite{kreutz2003dictionary,engan1999method,aharon2006img,chen2015dictionary,zayyani2015dictionary}.
Alternatively, polynomial-time algorithms that are guaranteed to reach a globally optimal solution  appear in~\cite{spielman2012exact,arora2013new}. 
 
In order to capture  multi-scale traits in natural signals, a method to construct multi-scale dictionaries was presented in \cite{mairal2008learning}. The multi-scale representation was obtained by using a quadtree decomposition of the learned dictionary. Alternatively, the work in \cite{ophir2011multi,yan2013nonlocal} applied dictionary learning in the domain of a fixed multi-scale operator (wavelets). In our approach we follow a different multi-scale strategy, based on  a pyramid decomposition, similar
to the Laplacian pyramid~\cite{burt1983laplacian}.

There exist dictionary learning approaches designed to couple multi-modal data. Monaci \textit{et al.} \cite{monaci2007learning}  proposed an  approach to learn basis functions representing  audio-visual structures. The approach, however, enforces synchrony between the different modalities. Alternatively, Yang \textit{et al.} \cite{yang2010image,yang2012coupled} considered  the problem of learning two dictionaries $D_x$ and $D_y$ for two families of signals $x,y$,  coupled by a mapping function $\mathcal{F}$ [with $y=\mathcal{F}(x)$]. The constraint was that the sparse representation of $x$ in $D_x$ is the same as that of $y$ in $D_y$. The  application targeted was image super-resolution, where $x$ (resp. $y$) is the low (resp. high)  resolution image.
The study in \cite{wang2012semi} followed a similar approach with the difference that the mapping function was applied to the sparse codes, i.e., $z_y=\mathcal{F}(z_x)$, rather than the signals. Jia \textit{et al.}  \cite{jia2010factorized} proposed  dictionary learning via the concept of group sparsity so as to couple the different views in human pose estimation. 
Our coupled dictionary learning method is designed to address the challenges of the targeted source separation application and as such, the model we consider to represent the correlated sources is fundamentally different from previous work. 
Moreover, we extend coupled dictionary learning to the multi-scale case and we provide a way to ignore certain noisy parts of the training signals (corresponding to cracks in our case).


\section{ Image Separation with Side Information} 
\label{sec:SourceSepSideInfor}
We denote by $x^{\text{ray}}_1$ and
$x^{\text{ray}}_2$  two vectorized X-ray image patches that we wish to separate from each other given  a mixture patch $m$, where~$m=x^{\text{ray}}_1+x^{\text{ray}}_2$.
Let $y_1$ and $y_2$ be the  co-located (visual)  image patches of the
front and  back of the painting. These patches play the role of \textit{side
information} that aids the separation. The use of side information has proven beneficial in various inverse problems
\cite{vaswani2010modified,mota2014comp,mota2014compressed,renna2014classification,scarlett2013compressed,zimos2016weighted,khajehnejad2009weighted}. In this work, we consider the signals $x^{\text{ray}}_1, x^{\text{ray}}_2, y_1, y_2\in\mathbb{R}^{n}$ to obey (superpositions of) sparse representations
in some dictionaries:
\begin{align}\label{eq:visual_model}
y_{1} &=\Psi^{c}z_{1c}  \nonumber\\
y_{2} &=\Psi^{c}z_{2c},
\end{align}
and
\begin{align}\label{eq:xray_model}
x^{\text{ray}}_{1} &=\Phi^{c}z_{1c} + \Phi v_1 \nonumber\\
x^{\text{ray}}_{2} &=\Phi^{c}z_{2c} + \Phi v_2,
\end{align}
where $z_{ic}\in \mathbb{R}^{\gamma\times1}$, with $\|z_{ic}\|_0=s_{z}\ll \gamma$
and $i=1,2$, denotes the sparse component  that is common
to the images in the visible and  the X-ray domain with respect to dictionaries
$\Psi^{c}, \Phi^{c}\in\mathbb{R}^{n\times\gamma}$, respectively. The parameter $s_z$ denotes the
sparsity of the vector $z_{ic}$. Moreover, $v_i\in \mathbb{R}^{d\times1}$, with $\|v_i\|_0=s_{v}\ll d_{}$
denotes the sparse innovation component of the X-ray image, obtained with
respect to the dictionary $\Phi\in\mathbb{R}^{n\times d}$. The common components express global features
and   structural characteristics that underlie both modalities. The innovation components capture parts
of the signal specific to the X-ray modality, that is, traces
of the wooden panel or even footprints of the vertical and horizontal wooden
slats attached to the back the painting.
We acknowledge the relation of our model with the sparse common component and innovations model that captures intra- and inter-signal correlation of physical signals in wireless sensor networks~\cite{renna2014classification,wakin2005recovery}. Our approach is however more generic, since we decompose the signals in learnt dictionaries rather than fixed canonical bases,  as in~\cite{wakin2005recovery}.

Given the proposed model and provided that the dictionaries $\Psi^c$,
$\Phi$, and $\Phi^c$ are known, the corresponding X-ray separation 
problem can be formulated as  
        \begin{align}\label{Eq:UnSolvedProb}
                \begin{array}{ll}
                        \underset{z_{1c}, z_{2c},v_1, v_{2}}{\text{minimize}}
& \|z_{1c}\|_1 + \|z_{2c}\|_1 + \|v_1\|_1 + \|v_2\|_1  \\
                        \:\:\quad\text{s.t.} & m = \Phi^{c}z_{1c} + \Phi^{c}z_{2c}
+ \Phi v_1 + \Phi v_2 \\
& y_{1} =\Psi^{c}z_{1c} \\
& y_{2} =\Psi^{c}z_{2c}
                \end{array}             
        \end{align}
where we  applied convex relaxation by replacing the $\ell_0$-pseudo norm with the $\ell_1$-norm, denoted as $\|\cdot\|_1$. Problem \eqref{Eq:UnSolvedProb} is under-determined, namely,  $v_1$ and $v_2$ cannot be distinguished due
to the symmetry in the constraints. A  solution to the unmixing problem
can be obtained when $v_1=v_2=v$,
which is formally written as   
        \begin{align}\label{Eq:Prob}
                \begin{array}{ll}
                        \underset{z_{1c}, z_{2c},v}{\text{minimize}}
& \|z_{1c}\|_1 + \|z_{2c}\|_1 + 2\|v\|_1  \\
                        \:\quad\text{s.t.} & m = \Phi^{c}z_{1c} + \Phi^{c}z_{2c}
+ 2\Phi v \\
& y_{1} =\Psi^{c}z_{1c} \\
& y_{2} =\Psi^{c}z_{2c}
                \end{array}             
        \end{align}
        Problem \eqref{Eq:Prob}  boils down to basis pursuit,
which is solved by convex optimization tools, e.g., \cite{BergFriedlander2008}.
The assumption $v_1=v_2$ is not only practical but is also motivated by the actual problem.
Since the paintings are mounted on
the same wooden panel, the sparse
components that decompose the X-ray images via the dictionary $\Phi$ are expected
to be the same.

\section{Coupled Dictionary Learning Algorithm}
\label{sec:coupledDicLearn}
In order to address the source separation with side information problem, we learn coupled dictionaries, $\Psi^c$, $\Phi^c$, $\Phi$, by using image
patches  sampled from 
visual and X-ray images of single-sided  panels, which 
do not suffer from superposition phenomena.
The images were registered using the algorithm described in~\cite{carreras06}.  Let $Y=[y_1,\dots, y_t],X=[x_1,\dots, x_t]\in\mathbb{R}^{n \times
t}$ represent a set of~$t$ co-located vectorized
visual and X-ray patches, each
containing $\sqrt{n}\times\sqrt{n}$
 pixels. As per our  model in \eqref{eq:visual_model} and \eqref{eq:xray_model}, the columns of $X$~and $Y$~are decomposed as
        \begin{subequations}\label{Eq:Model}
        \begin{align}
                Y &= \Psi^c Z  
                \label{Eq:ModelxRay}
                \\
                X &= \Phi^cZ + \Phi V\,,
                \label{Eq:ModelImages}
        \end{align}
        \end{subequations}
where we collect their common components into the columns of the matrix $Z=[z_1,\dots, z_t]
\in \mathbb{R}^{\gamma \times t}$ and their innovation components into the
columns of $V=[v_1,\dots, v_t]\in\mathbb{R}^{d \times{t}}$. 
We formulate the coupled dictionary learning problem as
        \begin{equation}\label{Eq:Problem}
                \begin{array}[t]{cl}
                        \underset{\begin{subarray}\\ \Psi^c , Z \\ \Phi^c,\Phi,V\end{subarray}}{\text{minimize}}
                        &
                                \frac{1}{2}\big\|Y - \Psi^c Z \big\|_F^2
                                +
                                \frac{1}{2}\big\|X - \Phi^c Z - \Phi V\big\|_F^2,
                        \\
                        \text{s.t.} &
                                 \big\|z_\tau\big\|_0 \leq s_{z},  
                          
                                \vspace{0.2cm}
                                \\&
                                \big\|v_\tau\big\|_0\leq s_{v}, \quad \forall
\tau=1,2,\dots ,t.
                \end{array}
        \end{equation}
Similar to related work \cite{aharon2006img,mairal2008learning,ophir2011multi}, we solve Problem~\eqref{Eq:Problem} by alternating between a sparse-coding step and a dictionary
update step. Particularly, given initial estimates for dictionaries $\Psi^c$,
$\Phi$, and $\Phi^c$---in line with prior work \cite{aharon2006img}  we use the overcomplete discrete cosine transform (DCT) for initialization---we iterate on~$k$ between a sparse-coding step:
\begin{align}
(Z^{k+1}, V^{k+1}) &=
                \begin{array}[t]{cl}
                        \underset{Z,V}{\text{arg\;min}} &
\frac{1}{2}
                        \Bigg\|
                                \begin{bmatrix}Y \\ X\end{bmatrix} 
                                -
                                \begin{bmatrix}
                                        {\Psi^c}^k &  0 \\
                                        {\Phi^c}^k & \Phi^k
                                \end{bmatrix}
                                \begin{bmatrix}
                                        Z  \\ V
                                \end{bmatrix} \vspace{0.2cm}
                        \Bigg\|_F^2,
                        \\
                        \text{s.t.} & \big\|z_\tau\big\|_0 \leq s_{z},     
                                \vspace{0.2cm}
                                \\&
                                \big\|v_\tau\big\|_0\leq s_{v}, \quad \forall
\tau=1,2,\dots
,t,
                \end{array}             
                \label{Eq:SolverWeights}
        \end{align}
which is performed to learn the sparse codes $Z,V$ having the dictionaries fixed, and a dictionary update step         
        \begin{align}
               ({\Psi^c}^{k+1},{\Phi^c}^{k+1},& \Phi^{k+1}) =
                \nonumber\\ 
                        &\arg\min_{\Psi^c, \Phi^c, \Phi} \frac{1}{2}
                        \Bigg\|
                                \begin{bmatrix}Y \\ X\end{bmatrix} 
                                -
                                \begin{bmatrix}
                                        {\Psi^c} &  0 \\
                                        {\Phi^c} & \Phi
                                \end{bmatrix}
                                \begin{bmatrix}
                                        Z^{k+1}  \\ V^{k+1}
                                \end{bmatrix} \vspace{0.2cm}
                        \Bigg\|_F^2.            
                \label{Eq:SolverDictionary}
        \end{align}
which updates the dictionaries given the calculated sparse codes. 
The algorithm iterates between these steps until no additional
iteration reduces the value of the cost function below a chosen threshold, or until a predetermined number of iterations is reached.  In what
follows, we provide details regarding the solution of the problem at each
stage.

\textbf{Sparse-coding step.} Problem~\eqref{Eq:SolverWeights}
        decomposes into~$t$ problems, each of which can be solved in parallel:
       \begin{align}
(z_\tau^{k+1}, v_\tau^{k+1}) &=
                \begin{array}[t]{cl}
                        \underset{z_\tau,v_\tau}{\text{arg\;min}} &
\frac{1}{2}
                        \Bigg\|
                                \begin{bmatrix}y_{\tau} \\ x_{\tau}\end{bmatrix}
                                -
                                \begin{bmatrix}
                                        {\Psi^c}^k &  0 \\
                                        {\Phi^c}^k & \Phi^k
                                \end{bmatrix}
                                \begin{bmatrix}
                                        z_\tau  \\ v_\tau
                                \end{bmatrix} \vspace{0.2cm}
                        \Bigg\|_F^2,
                        \\
                        \text{s.t.} & \big\|z_\tau\big\|_0 \leq s_{z},  
                                \vspace{0.2cm}
                                \\&
                                \big\|v_\tau\big\|_0\leq s_{v}.                \end{array}             
                \label{Eq:SolverWeights1}
        \end{align} 
To address each of the $t$ problems
in~\eqref{Eq:SolverWeights1}, we propose a greedy algorithm that constitutes
a modification of the   OMP method [see Algorithm~\ref{Alg:modifiedOMP}]. 
Our method adapts OMP~\cite{tropp2007signal} to solve:
\begin{align}
\begin{array}{cc}
\underset{w}{\text{minimize}} & \|b - \Theta w\|_2^2 \\
\text{s.t.} & \|w(\mathcal{I})\|_0\leq s_z, \\
 & \|w(\mathcal{J})\|_0\leq s_v,
\label{Eq:ompProb}
\end{array}
\end{align}
where $w(\mathcal{I})$ [resp., $w(\mathcal{J})$] denotes the components
of  vector $w\in \mathbb{R}^{(\gamma+d)\times1}$ indexed by the index set
$\mathcal{I}$
(resp., $\mathcal{J}$), with $\mathcal{I}\cup\mathcal{J}=\{1,2,\dots,\gamma+d\}$,
$\mathcal{I}\cap\mathcal{J}=\{\emptyset\}$. Each sub-problem
in \eqref{Eq:SolverWeights1} translates to \eqref{Eq:ompProb}  
by replacing: $b=\begin{bmatrix}y_\tau \\ x_\tau\end{bmatrix}$,
$\Theta=\begin{bmatrix}
        {\Psi^c}^k & 0 \\
        {\Phi^c}^k & \Phi^k
        \end{bmatrix}$,
and $w=\begin{bmatrix} z_\tau \\ v_\tau\end{bmatrix}$.  
     
\begin{algorithm}[t]   
    \caption{Modified Orthogonal Matching Pursuit (mOMP)}
    \algrenewcommand\algorithmicrequire{\textbf{Input:}}
    \label{Alg:modifiedOMP}
    \begin{algorithmic}[1]
    \small
    \Require 
    \State A vector $b\in\mathbb{R}^{(2n)\times1}$
    \State A matrix $\Theta\in\mathbb{R}^{(2n)\times{(\gamma+d)}}$,
    \State The indices $\mathcal{I}$, $\mathcal{J}$, with $\mathcal{I}\cup\mathcal{J}=\{1,2,\dots,\gamma+d\}$,
$\mathcal{I}\cap\mathcal{J}=\{\emptyset\}$
    \State The sparsity levels $s_z$, $s_v$ of vectors $z$ and $v$, respectively.
    \Statex
    \algrenewcommand\algorithmicrequire{\textbf{Output:}}          
     \Require $w$ (approximate) solution of \eqref{Eq:ompProb}.
 
                \Statex
                
                \algrenewcommand\algorithmicrequire{\textbf{Initialization}}
                \Require
                
\State Initialize the residual $r_0=b$.
\State Set the total sparsity of vector $w$ as $s_w=s_z+s_v$.
\State Set the counters for the sparsity of $z$ and $v$ to $\ell_z=0$, $\ell_v=0$.
\State Initialize the set of non-zero elements of $w$ to $\Omega=\emptyset$.   
                 
                \Statex                                                 
                    
                \algrenewcommand\algorithmicrequire{\textbf{Algorithm}}
                \Require
                                    
\For{$i = 1,2,\dots, s_w$}                          
\State Sort the indices $\zeta=\{1,2,\dots,\gamma+d\}$, corresponding to
the $\theta_\zeta$ columns of $\Theta$,  such that $|\langle r_{i-1}, \theta_{\zeta}
\rangle|$ are in descending order (where $\langle \alpha,\beta\rangle$ denotes the inner product of the vectors  $\alpha,\beta$). Then, put the ordered indices in the vector
 $q_{i}$.
\State Set $\mathcal{G}=\emptyset$ and auxiliary iterator $\texttt{iter}=0$.
           \While{$\mathcal{G}=\emptyset$}
                \State{\texttt{iter}=\texttt{iter}+1}           
                \State Find the index of the $\Theta$ matrix that corresponds
to the value of \texttt{iter}: {$\kappa = q_i[\texttt{iter}]$.} 
                \If  {$\kappa\in\mathcal{I}$ \text{AND} $\ell_z<s_z$}
                 \State {Set $\mathcal{G}=\kappa$ and increase: $\ell_z=\ell_z +1$.}
                 \Else
                 \If{$\kappa\in\mathcal{J}$ \text{AND} $\ell_v<s_v$}
                \State {Set $\mathcal{G}=\kappa$ and increase: $\ell_v=\ell_v +1$.}
                 \EndIf
                 \EndIf
                 \EndWhile
                 \State Update the set of non-zero elements of $w$, i.e.,
$\Omega_{i} = \Omega_{i-1}\cup~\{\kappa\}$, and the matrix of chosen atoms:
$\Theta_i=[\Theta_{i-1}\quad \theta_{\kappa}]$.
                 \State Solve: $w_i=\arg\min_w\|b-\Theta_i
w\|_2$.
                 \State Calculate the new residual: $r_i=b-\Theta_i w_i$.

                 \EndFor
    \end{algorithmic}
  \end{algorithm}
\textbf{Dictionary update step.} Problem~\eqref{Eq:SolverDictionary}
can be written as
        \begin{equation}\label{Eq:SolverDictionary1}
                \begin{array}[t]{ll}
                        \underset{\Psi^c,\overline{\Phi}}{\text{minimize}}
                        &
                        \frac{1}{2}\Big\|Y - \Psi^c Z^{k+1}\Big\|_F^2
                        +
                        \frac{1}{2}\Big\|X - \overline{\Phi}\:\overline{V}^{k+1}\Big\|_F^2
                        ,
                \end{array}
        \end{equation}
where $\overline{\Phi} = \begin{bmatrix}\Phi^c & \Phi\end{bmatrix}$ and $\overline{V}^{k+1} = \begin{bmatrix}Z^{k+1} \\ V^{k+1}\end{bmatrix}$. Problem~\eqref{Eq:SolverDictionary1} decouples into two (independent)
problems, that is,
                \begin{equation}
                \begin{array}[t]{ll}
                        \underset{\Psi^c}{\text{minimize}}
                        &
                        \frac{1}{2}\Big\|Y - \Psi^c Z^{k+1}\Big\|_F^2
                \end{array}
                \label{eq:MODproblem1} 
                \end{equation}
and
                \begin{equation}
                \begin{array}[t]{ll}
                        \underset{\overline{\Phi}}{\text{minimize}}
                        &
                        
                        \frac{1}{2}\Big\|X - \overline{\Phi}\:\overline{V}^{k+1}\Big\|_F^2
                        .
                \end{array}
                \label{eq:MODproblem2}
                \end{equation}
Provided that $Z^{k+1}$ and $\overline{V}^{k+1}$ are full row-rank, each of these problems
has a closed-form solution, namely, 
$${\Psi^{c}}^{k+1}=Y{Z^{k+1}}^T\Big({Z^{k+1}}{Z^{k+1}}^T\Big)^{-1}$$
and
$$\overline{\Phi}^{k+1}=X{\overline{V}^{k+1}}^T\Big({\overline{V}^{k+1}}{\overline{V}^{k+1}}^T\Big)^{-1}.$$
When $Z^{k+1}$ and $\overline{V}^{k+1}$ are rank-deficient,~\eqref{eq:MODproblem1} and \eqref{eq:MODproblem2} have multiple solutions, from which we select
the one with minimal Frobenius norm.  This is done by taking a thin singular value decomposition of~$Z^{k+1}=G_{z^{k+1}}\Sigma_{z^{k+1}}U_{z^{k+1}}^T$ and~$\overline{V}^{k+1}=G_{\overline{v}^{k+1}}\Sigma_{\overline{v}^{k+1}}U_{\overline{v}^{k+1}}^T$, and calculating  
$$
{\Psi^{c}}^{k+1}= YU_{z^{k+1}}\Sigma_{z^{k+1}}^{-1} G_{z^{k+1}}^T
$$
and 
$$
\overline{\Phi}^{k+1}= XU_{\overline{v}^{k+1}}\Sigma_{\overline{v}^{k+1}}^{-1} G_{\overline{v}^{k+1}}^T.
$$ 
\section{Weighted Coupled Dictionary Learning}
\label{sec:weightedcoupledDicLearn}
Visual and X-ray images of paintings  contain a high number of pixels
that depict cracks. These are fine patterns of dense cracking formed within the materials. When taking into
account these pixels, the learned dictionaries  comprise
atoms that correspond to high frequency components. As a consequence, 
the reconstructed  images are contaminated by high frequency
noise. In order to improve the separation performance, our objective  is to obtain dictionaries that ignore pixels  representing  cracks. To identify  such pixels, we  generate binary masks  identifying the location of  cracks by applying our method in \cite{cornelis2012crack}. Each sampled image patch may contain a variable number of crack pixels, meaning that each column of the data matrix contains a different number of meaningful entries.
To address this issue, we introduce a weighting scheme that adds a
weight of 0 or 1 to the pixels that do or do not correspond to cracks, respectively.
These  crack-induced weights  are included using a Hadamard
product, namely, our model in \eqref{Eq:Model} is
 modified to 
       \begin{subequations}\label{Eq:WeightedModel}
        \begin{align}
                Y\odot\Lambda &= (\Psi^c Z)\odot\Lambda  
                \label{Eq:WeightedModelxRay}
                \\
                X\odot\Lambda &= (\Phi^cZ + \Phi V)\odot\Lambda.
                \label{Eq:WeightedModelImages}
        \end{align}
        \end{subequations}
where the matrix $\Lambda$ has exactly the same dimensions as $X$ and $Y$ and its entries are 0 or 1 depending on whether a pixel is part of a crack or not, respectively. We now formulate the weighted coupled dictionary learning problem
as
        \begin{equation}\label{Eq:WeightedProblem}
                \begin{array}[t]{cl}
                        \underset{\begin{subarray}\\ \Psi^c , Z \\ \Phi^c,\Phi,V\end{subarray}}{\text{minimize}}
                        &
                                \frac{1}{2}\big\|(Y - \Psi^c Z)\odot\Lambda
\big\|_F^2
                                \\&\qquad\qquad\qquad+ \frac{1}{2}\big\|(X - \Phi^c Z - \Phi V)\odot\Lambda\big\|_F^2,
                        \\
                        \text{s.t.} &
                                 \big\|z_\tau\big\|_0 \leq s_{z},  
                          
                                \vspace{0.2cm}
                                \\&
                                \big\|v_\tau\big\|_0\leq s_{v}, \quad \forall
\tau=1,2,\dots t.
                \end{array}
        \end{equation}
        Similar to \eqref{Eq:Problem}, the solution for \eqref{Eq:WeightedProblem} is obtained by  alternating between a sparse-coding and a dictionary update step. 

        \textbf{Sparse-coding step.} Similar to~\eqref{Eq:SolverWeights1}, the sparse-coding problem decomposes into~$t$ problems that can be solved in parallel:
        \begin{align}\label{Eq:NewSolverWeights1}
                (z_\tau^{k+1}, v_\tau^{k+1}) 
                = 
                &\underset{z_\tau, v_\tau}{\text{arg\;min}}
                \,\,\,
                \frac{1}{2}
                        \Bigg\|
                                \begin{bmatrix}y_\tau \\ x_\tau\end{bmatrix} 
                                \odot \begin{bmatrix}\lambda_\tau \\ \lambda_\tau\end{bmatrix}
                                \nonumber\\&\qquad -
                                \Big(\begin{bmatrix}
                                        {\Psi^c}^k & 0 \\
                                        {\Phi^c}^k & \Phi^k
                                \end{bmatrix}
                                \odot \Big(\begin{bmatrix}\lambda_\tau \\ \lambda_\tau\end{bmatrix}
                                \boldsymbol{1}^T\Big)\Big)
                                \begin{bmatrix}
                                        z_\tau \\ v_\tau
                                \end{bmatrix}
                        \Bigg\|_2^2\nonumber\\
                        &\quad \text{s.t.} \
                        \:\|z_\tau\|_0 \leq s_{z}, \nonumber\\       
                             & \qquad\:\:\:\|v_\tau\|_0\leq s_{v}, \quad \forall
\tau=1,2,\dots t,
        \end{align} 
where we used $\lambda_\tau$ to represent column~$\tau$
of $\Lambda$ and $\boldsymbol{1}^T$ to denote a row-vector of ones with dimension
equal to $\gamma+d$. To address each of the $t$ sub-problems in~\eqref{Eq:NewSolverWeights1},
we use the mOMP algorithm described in Algorithm \ref{Alg:modifiedOMP}, as each sub-problem in \eqref{Eq:NewSolverWeights1}
reduces  to \eqref{Eq:ompProb} by replacing: $b=\begin{bmatrix}y_\tau \\ x_\tau\end{bmatrix}\odot
\begin{bmatrix}\lambda_\tau \\ \lambda_\tau\end{bmatrix}$,
$\Theta=
                                \begin{bmatrix}
                                        {\Psi^c}^k & 0 \\
                                        {\Phi^c}^k & \Phi^k
                                \end{bmatrix}
                                \odot \Big(\begin{bmatrix}\lambda_\tau \\ \lambda_\tau\end{bmatrix}
                                \boldsymbol{1}^T\Big)$,
and $w=\begin{bmatrix}
                                        z_\tau \\ v_\tau
                                \end{bmatrix}$. 
          
        \textbf{Dictionary update step.} The dictionary update problem is now written as
        \begin{equation}\label{Eq:SolverWeightedDictionary1}
                \begin{array}[t]{ll}
                        \underset{\Psi^c,\overline{\Phi}}{\text{minimize}}
                        &
                        \frac{1}{2}\Big\|Y\odot\Lambda - (\Psi^c Z^{k+1})\odot\Lambda\Big\|_F^2
                        \\&\qquad\quad+
                        \frac{1}{2}\Big\|X\odot\Lambda- (\overline{\Phi}\:\overline{V}^{k+1})\odot\Lambda\Big\|_F^2
                        ,
                \end{array}
        \end{equation}
        and it decouples into:
        \begin{subequations}
        \begin{align}
                  &\underset{\Psi^c}{\text{minimize}}
                        \frac{1}{2}\Big\|Y\odot\Lambda - (\Psi^c Z^{k+1})\odot\Lambda\Big\|_F^2
                        \label{Eq:DictUpdateProblem1}
                        \\
                &\underset{\overline{\Phi}}{\text{minimize}}
                  \frac{1}{2}\Big\|X\odot\Lambda- (\overline{\Phi}\:\overline{V}^{k+1})\odot\Lambda\Big\|_F^2
                        .
                \label{Eq:DictUpdateProblem2}
                \end{align}
        \end{subequations}
We present only the solution
of the first problem since the solution of the other follows the
same logic. Specifically, we express the Frobenius norm in \eqref{Eq:DictUpdateProblem1}
as the sum of $t$ $\ell_2$-norm terms, each corresponding to a vectorized
training patch
\begin{equation}
\sum_{\tau=1}^t 
\|y_\tau \odot \lambda_\tau - ({\Psi^c}  z_\tau )\odot \lambda_\tau\|_2^2.
\label{Eq:weightedFrobSumNotation}
\end{equation}
By replacing the Hadamard product with multiplication with a diagonal matrix
$\Delta_\tau=\text{diag}(\lambda_\tau)$, \eqref{Eq:weightedFrobSumNotation} can
be written as
\begin{equation}
\sum_{\tau=1}^t \|\Delta_\tau  y_\tau - \Delta_\tau  {\Psi^c}  z_\tau \|_2^2.
\label{Eq:mulL2normSumNotation}
\end{equation}
To minimize the expression in \eqref{Eq:mulL2normSumNotation}, we take the
derivative with respect to the dictionary ${\Psi^c}$ and  set it to zero:
\begin{align}
&\frac{\partial}{\partial\Psi^c} \sum_{\tau=1}^t \|\Delta_\tau  y_\tau - \Delta_\tau {\Psi^c}  z_\tau \|_2^2 = 0 \nonumber\\
\Longrightarrow&\sum_{\tau=1}^t \frac{\partial}{\partial\Psi^c}\Big[(\Delta_\tau
y_\tau - \Delta_\tau {\Psi^c} z_\tau)^T (\Delta_\tau y_\tau -
\Delta_\tau {\Psi^c}  z_\tau) \Big] = 0\nonumber\\
\Longrightarrow&2\sum_{\tau=1}^t \frac{\partial}{\partial\Psi^c}y_\tau^T\Delta_\tau^T\Delta_\tau{\Psi^c}
z_\tau =\sum_{\tau=1}^t \frac{\partial}{\partial\Psi^c} z_\tau^T  {{\Psi^c}}^T\Delta_\tau^T\Delta_\tau{\Psi^c}
 z_\tau \nonumber\\
\Longrightarrow&\sum_{\tau=1}^t \Big(\Delta_\tau^T\Delta_\tau y_\tau\boldsymbol{1}^T\Big)\odot(\boldsymbol{1}
z_\tau^T)\nonumber\\
&\qquad
=\sum_{\tau=1}^t \Big({{\Psi^c}}  z_\tau z_\tau^T\Big)\odot((\lambda_\tau\odot\lambda_\tau)\boldsymbol{1}^T).
\label{Eq:DerivativeL2}
\end{align}
Before proceeding with the method to solve \eqref{Eq:DerivativeL2}, we recall that the entries of  $\lambda_\tau$ are either 0 or 1. To avoid dividing by zero when solving \eqref{Eq:DerivativeL2}, we have to update the rows of the dictionary matrix one-by-one. 
Specifically,
for each row $i$ of ${\Psi^c}$, we consider the matrix $A_i
= \sum_{\tau \in S_i} z_\tau z_\tau^T$, where $S_i$ is the support\footnote{The support
$S_i$ is defined by the indices where the $i$-th row of $\Lambda$ is equal
to 1.} of the $i$-th row of $\Lambda$, and $z_\tau$ is the $\tau$-th column of $Z$. We also create  a vector $c_i = \sum_{\tau \in S_i} Y(i,\tau)z_\tau$, where $Y(i,\tau)$
is the $(i,\tau)$-th entry of $Y$. Provided that $A_i$ is invertible, the $i$-th row of ${\Psi^c}$ (which we denote
by the row-vector ${\psi^c_i}$) will  be given by
\begin{align}
\label{eq:updateDicCol}
{\psi^c_i} = c_i A_i^{-1}. 
\end{align}
If each $z_\tau$ is drawn randomly,  $A_i$ is invertible with probability~1 whenever the cardinality of $S_i$ is at least equal to the number of columns of $\Psi_c$. Although in practice each $z_\tau$ is not  randomly drawn, we still obtain an invertible   $A_i$ by guaranteeing that the number of training samples is large enough.    

\begin{figure}[t]
        \centering
\includegraphics[width=0.38\textwidth]{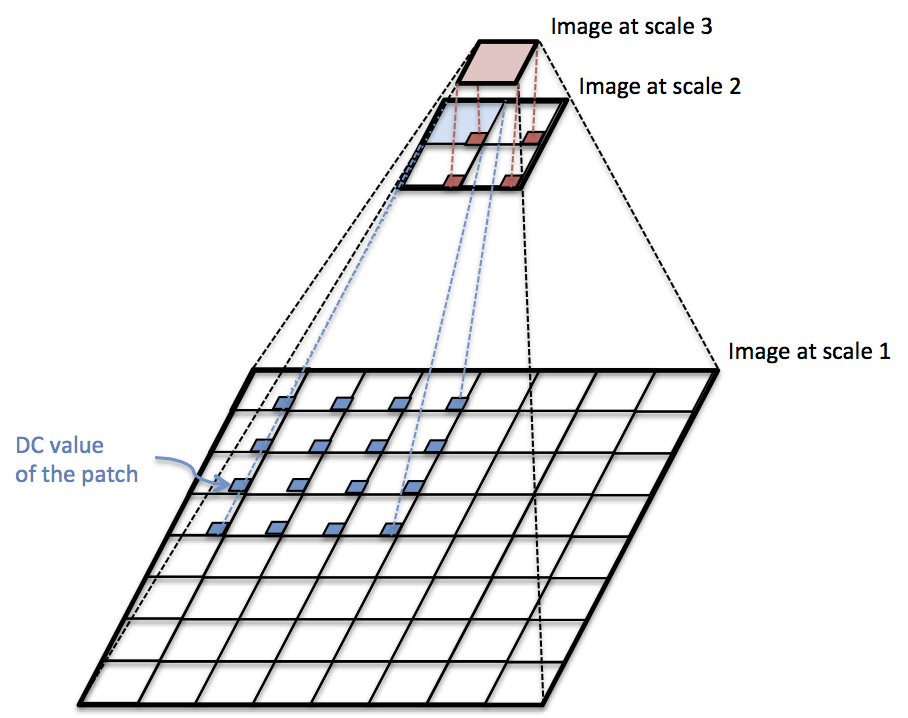}
        \caption{Schema of a 3-scale pyramid decomposition in the proposed
multi-scale dictionary learning and source separation approaches.}
        \label{Fig:pyramidSchema}
\end{figure}

\begin{figure}[t]
        \centering
\includegraphics[width=0.5\textwidth]{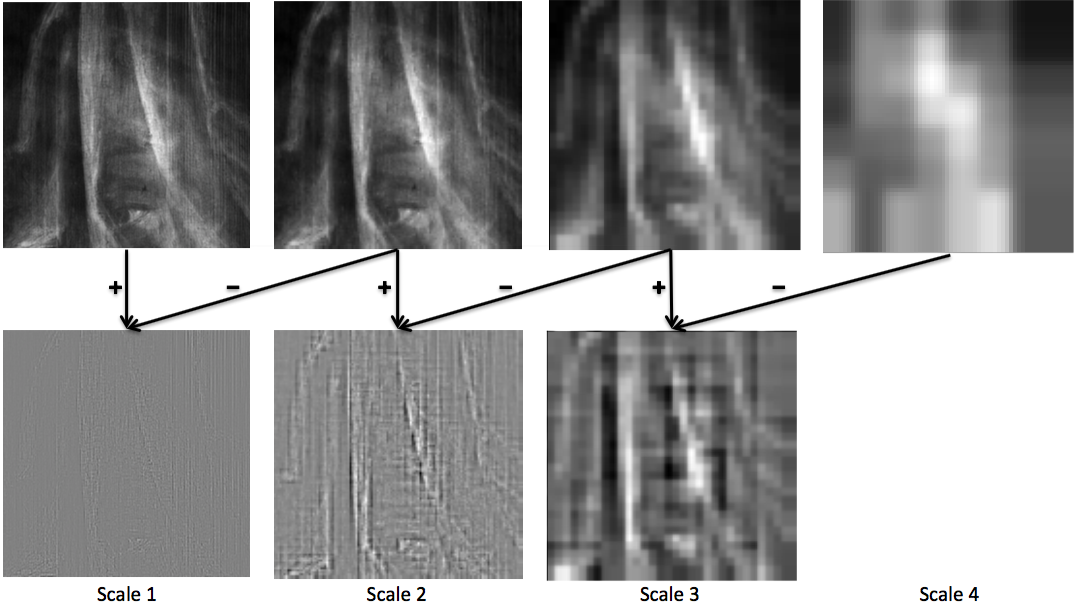}
        \caption{Example of a 4-scale pyramid decomposition of a mixed X-ray
image. The original image resolution is $1024\times 1024$ pixels. At scale
1, the image is split into non-overlapping patches of $8\times8$ pixels and
the DC value of every patch is extracted, thereby generating the high-pass
component. The aggregated DC values compose the low-pass  component
at scale 2, the resolution of which is $128\times128$ pixels. Dividing this
component into non-overlapping patches of $4\times4$ pixels and extracting
the DC value from every patch yields the high-pass band in scale 2. The procedure is repeated until finally the low-pass band at scale 4 has a resolution of $8\times8$
pixels.}
        \label{Fig:LapPyramidExm}
\end{figure}

\section{Single- And Multi-Scale Separation Methods}
\label{sec:MultScAppr}
\subsection{Single-Scale Approach}
\label{sec:SingScaleAppr} 
Given the trained coupled dictionaries, the
 source separation method described in Section \ref{sec:SourceSepSideInfor}
is applied locally, per overlapping patch of the X-ray image. Let the corresponding patches from the mixed
X-ray and the two corresponding visual images be denoted as $m^{\boldsymbol{u}}$, $y_1^{\boldsymbol{u}}$,
and  $y_2^{\boldsymbol{u}}$, respectively. Each patch contains $\sqrt{n}\times
\sqrt{n}$ pixels and has top-left coordinates 
$$
\boldsymbol{u} = (\epsilon\cdot u_1, \epsilon\cdot u_2), \quad 0\leq u_1<\left\lfloor
\frac{H}{\epsilon} \right\rfloor, \quad 1\leq u_2<\left\lfloor
\frac{W}{\epsilon} \right\rfloor,
$$     
where $\epsilon\in\mathbb{Z}_{+}$, $1\leq\epsilon<\sqrt{n}$ is the overlap
step-size, $\left\lfloor \bullet \right\rfloor$ is the floor function, and
$H, W$ are the image height and width, respectively.  Prior to separation,
the DC  value is removed from the pixels in each overlapping patch
and the residual  values are vectorized. The solution of Problem  \eqref{Eq:Prob} yields the  sparse components $z_{1c}^{\boldsymbol{u}},
z_{2c}^{\boldsymbol{u}}, \text{and} \ v^{\boldsymbol{u}}$ corresponding to
the  patch with coordinates $\boldsymbol{u}$. The texture  of each separated
 patch is then reconstructed following the model in \eqref{eq:xray_model},
that is,
$x_1^{\boldsymbol{u}} = \Phi^{c}z_{1c}^{\boldsymbol{u}} + \Phi v^{\boldsymbol{u}}$
and $x_2^{\boldsymbol{u}} = \Phi^{c}z_{2c}^{\boldsymbol{u}} + \Phi v^{\boldsymbol{u}}$. In certain cases, the $v$ component may capture parts of the actual content; for example, vertical brush strokes can be misinterpreted as  the wood texture of the panel. In this case, we can choose to skip the  $v$ component; namely, we can reconstruct
the texture of the X-ray patches as $x_1^{\boldsymbol{u}} = \Phi^{c}z_{1c}^{\boldsymbol{u}}
$ and $x_2^{\boldsymbol{u}} = \Phi^{c}z_{2c}^{\boldsymbol{u}}$. The DC values
are weighted according to the DC values of the co-located patches in the visual images and then added back to the corresponding separated X-ray patches.
Finally, the pixels in each separated X-ray are recovered as the average
of the co-located pixels in each overlapping patch. 

\subsection{Multi-Scale Approach}
\label{sec:MultiScaleSubSection} 
Due to the  restricted patch size in comparison
to the high resolution of
the X-ray image, the DC values of all patches carry a considerable amount
of the total image energy. In the single-scale approach, these DC values
are common to the two separated X-rays, thereby leading to  poor separation. To address this issue, we devise a multi-scale image separation approach. In contrast with the techniques in \cite{yan2013nonlocal,ophir2011multi,mairal2008learning},
the proposed multi-scale approach performs a pyramid decomposition  of the mixed X-ray and
visual images, that is, the images are recursively decomposed into low-pass
and high-pass bands. The  decompositions  at scale $l=\{1, 2,\dots,L\}$ are
constructed as follows. The images at scale $l$---where we use the notation
$M_l, Y_{1,l}, Y_{2,l}$, to refer to the mixed X-ray and the
two visuals, respectively---are divided into overlapping patches  $m_l^{\boldsymbol{u}_l},
y_{1,l}^{\boldsymbol{u}_l}, \:\text{and}\: y_{2,l}^{\boldsymbol{u}_l}$, each
of size $\sqrt{n_l}\times\sqrt{n_l}$ pixels. Each patch has top-left coordinates
$$
\boldsymbol{u}_l = (\epsilon_l\cdot u_{1,l}, \epsilon_l\cdot u_{2,l}), \:
0\leq u_{1,l}<\left\lfloor
\frac{H_l}{\epsilon_l} \right\rfloor, \: 0\leq u_{2,l}<\left\lfloor
\frac{W_l}{\epsilon_l} \right\rfloor,
$$
where $\epsilon_l\in\mathbb{Z}_+, \: 1\leq \epsilon_l<\sqrt{n_l}$
is the overlap step-size, and  $H_l, W_l$ are the height and width  of the
image decomposition at
scale $l$. The DC value is extracted from  each patch, thereby constructing
the high frequency  band of the image  at scale $l$. The aggregated
DC values comprise the low-pass component of the image, the resolution of
which  is $\left\lfloor\frac{H_l}{\epsilon_l} \right\rfloor\times\left\lfloor
\frac{W_l}{\epsilon_l} \right\rfloor$ pixels. The low-pass component is then
decomposed further at the subsequent scale ($l+1$). The pyramid decomposition
is schematically sketched in Fig. \ref{Fig:pyramidSchema} and exemplified
in Fig. \ref{Fig:LapPyramidExm}.   

The texture of the mixed X-ray image  at scale $l$ is separated patch-by-patch
by solving Problem  \eqref{Eq:Prob}. 
The texture  of
each separated  patch is then reconstructed as,
$x_{1,l}^{\boldsymbol{u}_l} = \Phi^{c}_lz_{1c,l}^{\boldsymbol{u}_l} + \Phi_l
v_l^{\boldsymbol{u}_l}$
and $x_{2,l}^{\boldsymbol{u}_l} = \Phi^{c}_lz_{2c,l}^{\boldsymbol{u}_l} +
\Phi_l
v_l^{\boldsymbol{u}_l}$;
or alternatively,  as $x_{1,l}^{\boldsymbol{u}_l} = \Phi^{c}_lz_{1c,l}^{\boldsymbol{u}_l}
$ and $x_{2,l}^{\boldsymbol{u}_l} = \Phi_l^{c}z_{2c, l}^{\boldsymbol{u}_l}$.
Note that the dictionary learning process can be applied per scale, yielding
a triple of coupled dictionaries $(\Psi^c_l,\Phi^c_l,\Phi_l)$ per scale $l$.
In practice, due to  lack of training data in the higher scales, dictionaries
are  learned only from the low-scale decompositions and then copied
to the higher scales.
  
The separated X-ray images are finally reconstructed by following the reverse
operation: descending the pyramid, the separated component
at the coarser scale is up-sampled and added to the separated component of
the finer scales. 
  
\section{Experiments}
\label{Sec:ExpSection}
\subsection{Experiments with Synthetic Data}        
As in  \cite{engan1999method,aharon2006img}, we first evaluate
the performance of our  coupled dictionary learning algorithm---described in Section~\ref{sec:coupledDicLearn}---and our source separation with side information method (see Section \ref{sec:SourceSepSideInfor}) using synthetic data. Firstly, we generate synthetic signals, $x$, $y$, according to model \eqref{eq:visual_model}, \eqref{eq:xray_model}, using random dictionaries and then, given the data, we  assess whether the algorithm recovers the original dictionaries. 
The random dictionaries $\Psi^c$, $\Phi$, and $\Phi^c$ of size $40\times{60}$ contain  entries drawn
    from the standard normal distribution  and their columns are normalized to have unit $\ell_2$-norm.
Given the  dictionaries,~$t=1500$ sparse vectors $Z$ and $V$ were
  produced, each with dimension~$\gamma=d=60$.  The column-vectors $z_{\tau}$
and $v_{\tau}$, $\tau=1,2,\dots,t$, comprised of respectively $s_z=2$ and $s_v=3$
non-zero  coefficients distributed uniformly and placed   in random and independent
locations. Combining the dictionaries and the sparse vectors according to
the model in \eqref{Eq:Model} yields the correlated data signals $X$ and
$Y$, to which white Gaussian noise with a varying signal-to-noise ratio (SNR)
has been added.


\begin{table}[t]
\caption{Dictionary Identifiability  of the Proposed 
Algorithm Based on Synthetic Data, Expressed in Terms of the Percentage of
Retrieved Atoms for the  Dictionaries in Model \eqref{Eq:Model}.} 
\label{tab:DictionaryIdentifiability}
\centering 
\tabcolsep=0.05cm
\footnotesize
\begin{tabular}{c|c| c|c| c|c| c|c} 
\hline\hline
SNR [dB] & $\infty$ & 40 & 35 & 30 & 25  & 20 & 15 \\[0.1ex]
\hline 
$\Psi^c$ & 96\% & 95.18\% &  95.38\%& 95.65\%& 95.20\%& 90.42\%&  12.53\% \\ 
$\Phi^c$  & 96.78\% & 95.97\% & 96.53\% & 96.50\% & 95.48\% & 74.35\% & 0.27\%\\
\!\!$\Phi$  & 92.95\% & 91.90\% & 91.73\% & 91.27\% & 91.50\% & 88.25\% & 3.07\%\\
\hline\hline 
\end{tabular}
\end{table}

\begin{table}[t]
\caption{Reconstruction error of the proposed Source Separation With Side Information Method based on Synthetic Data.} 
\label{tab:SourceSeparation}
\centering 
\tabcolsep=0.05cm
\footnotesize
\begin{tabular}{c|c| c|c| c|c| c|c} 
\hline\hline
SNR [dB] & $\infty$ & 40 & 35 & 30 & 25  & 20 & 15 \\[0.1ex]
\hline 
$x_1$ & $1.66\times 10^{-5}$ & 0.0094 &  0.0307& 0.0734& 0.1215& 0.2125&  0.5854
\\ 
$x_2$  & $1.43\times 10^{-5}$ & 0.0097 & 0.0222 & 0.0818 & 0.0857 & 0.2516 & 0.2781\\
\hline\hline 
\end{tabular}
\end{table}

\begin{figure}[t]
 \centering
  \subfigure[Visual]{
    \label{fig:visualexample:a}
    \includegraphics[width=0.15\textwidth]{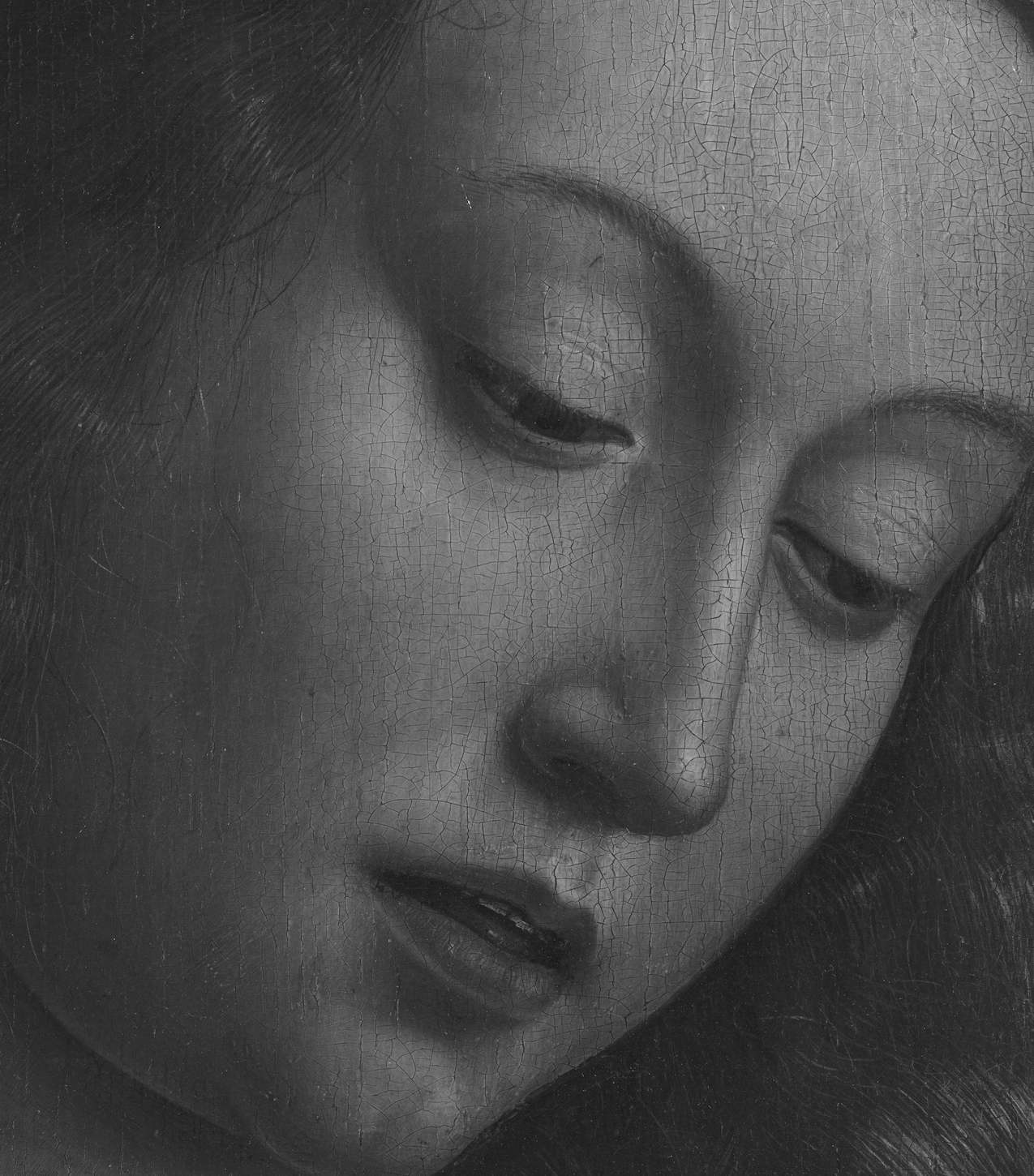}}
  \subfigure[X-ray \tiny\copyright KIK-IRPA]{
    \label{fig:visualexample:b}
    \includegraphics[width=0.15\textwidth]{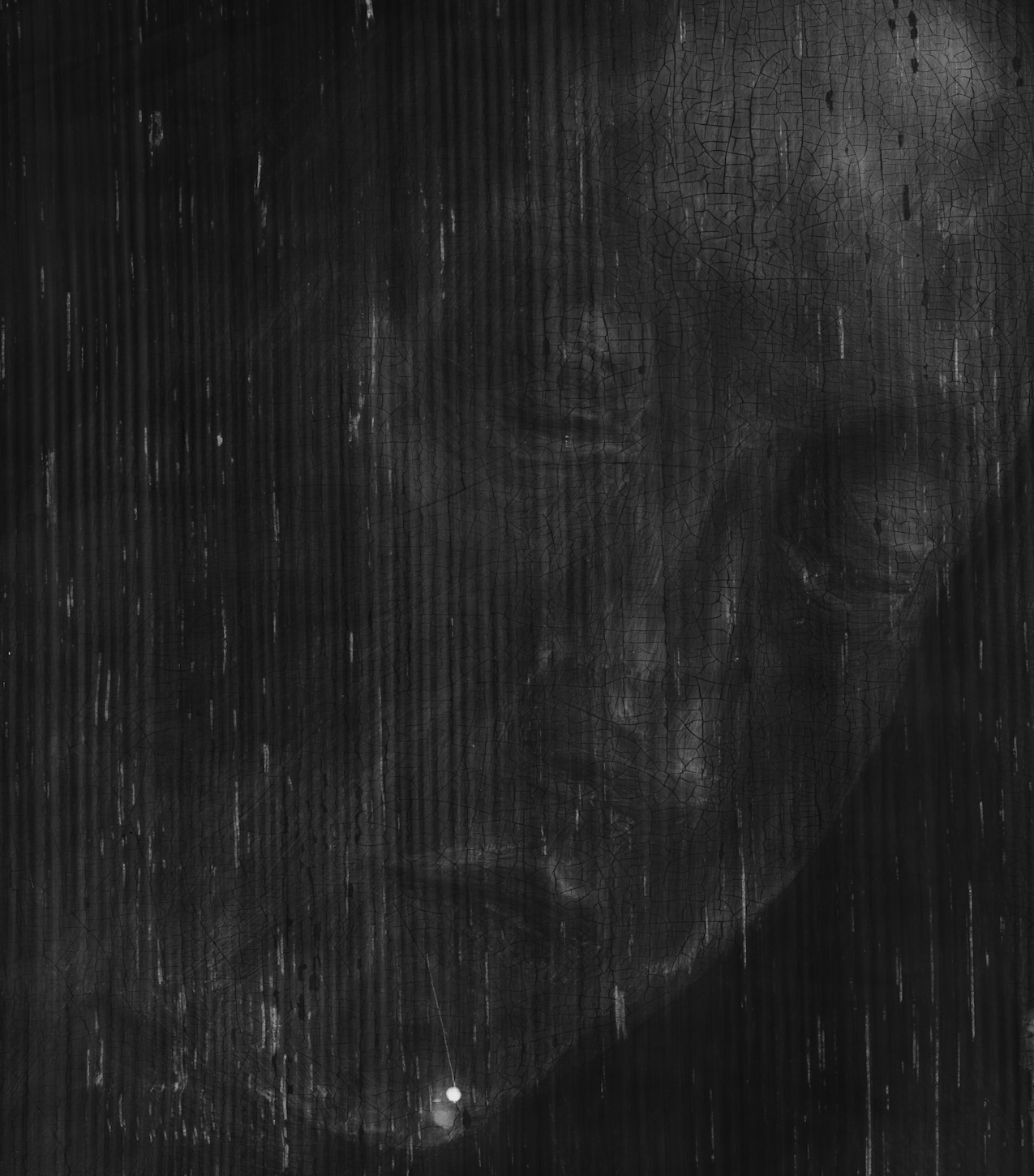}}
  \subfigure[Crack mask]{
    \label{fig:visualexample:c}
    \includegraphics[width=0.15\textwidth]{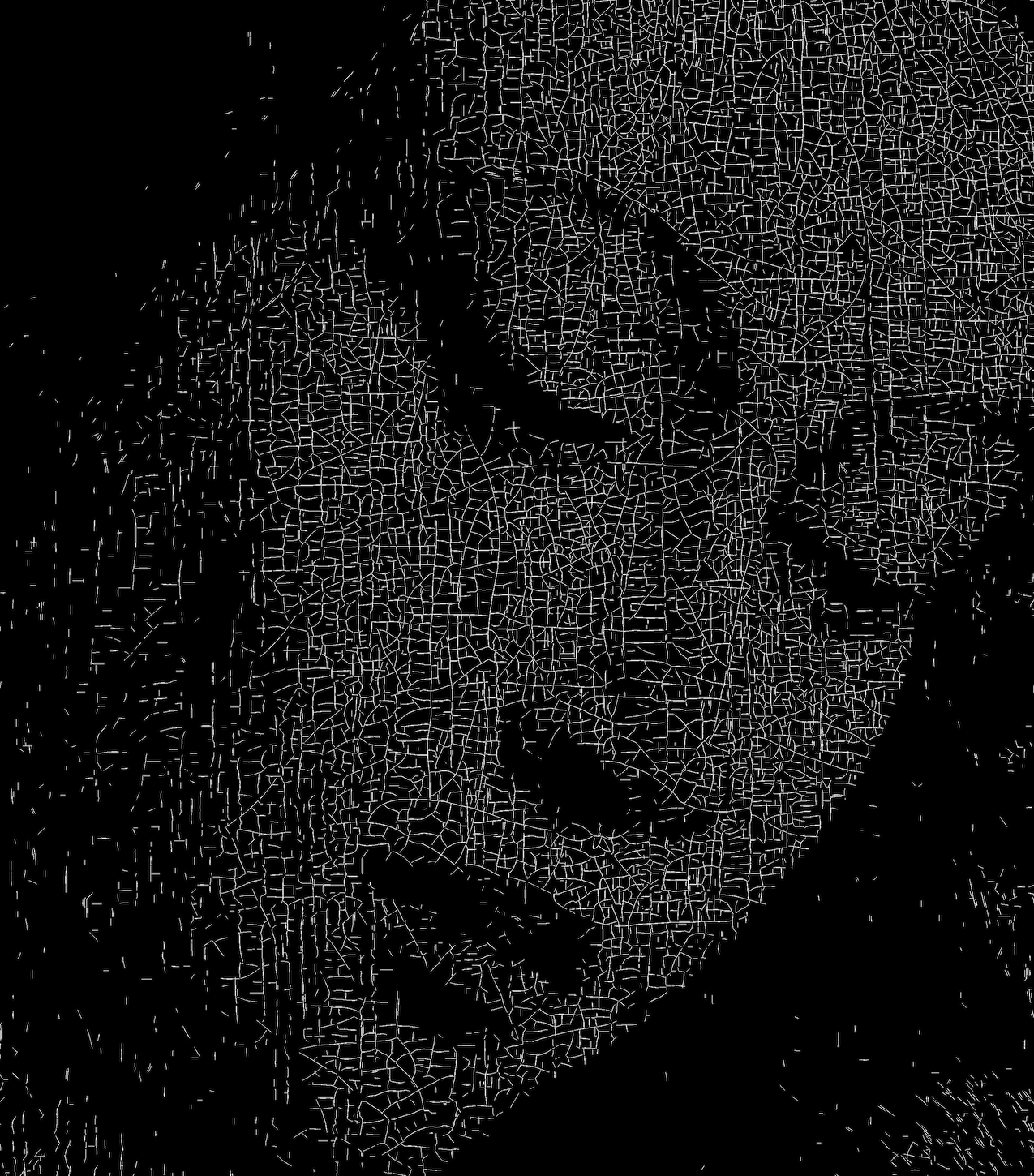}}
  \subfigure[Visual]{
    \label{fig:visualexample:a}
    \includegraphics[width=0.15\textwidth]{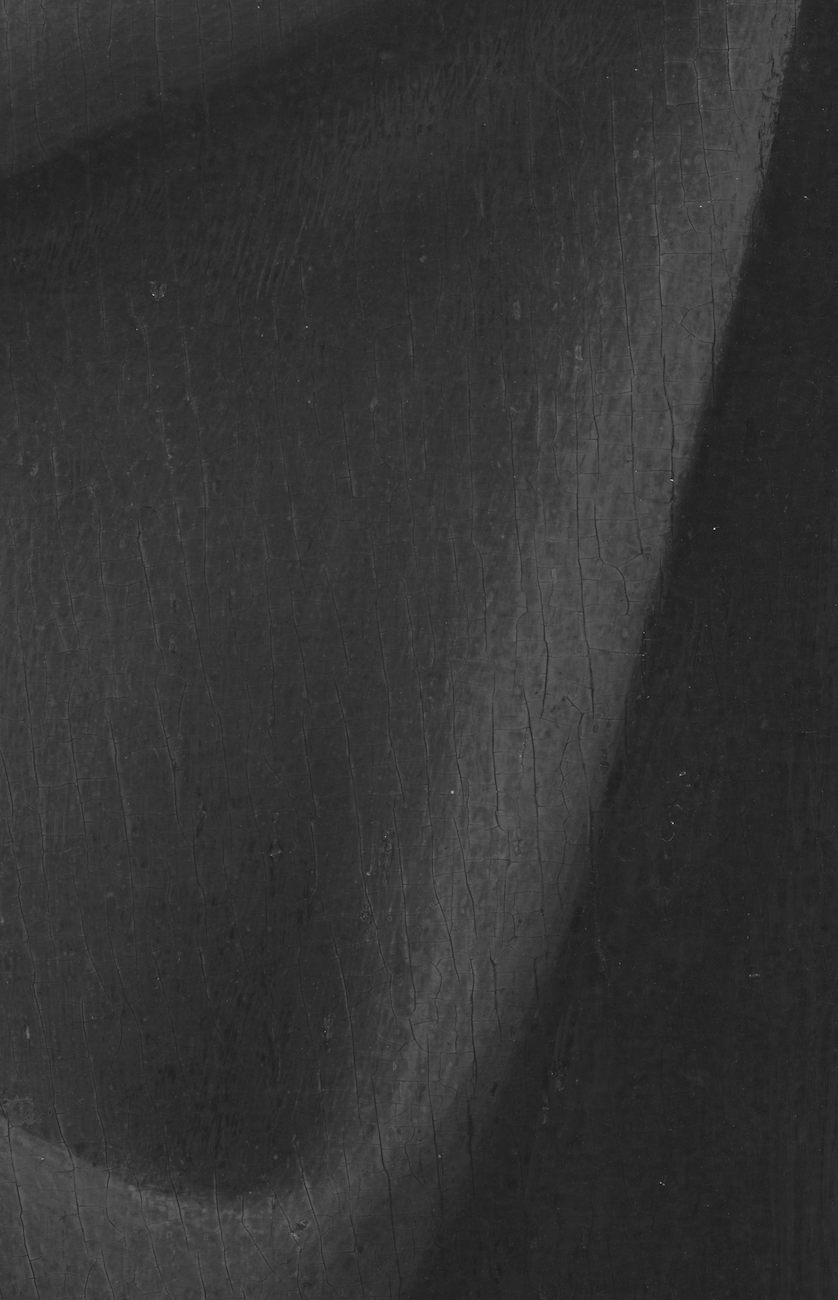}}
  \subfigure[X-ray \tiny\copyright KIK-IRPA]{
    \label{fig:visualexample:b}
    \includegraphics[width=0.15\textwidth]{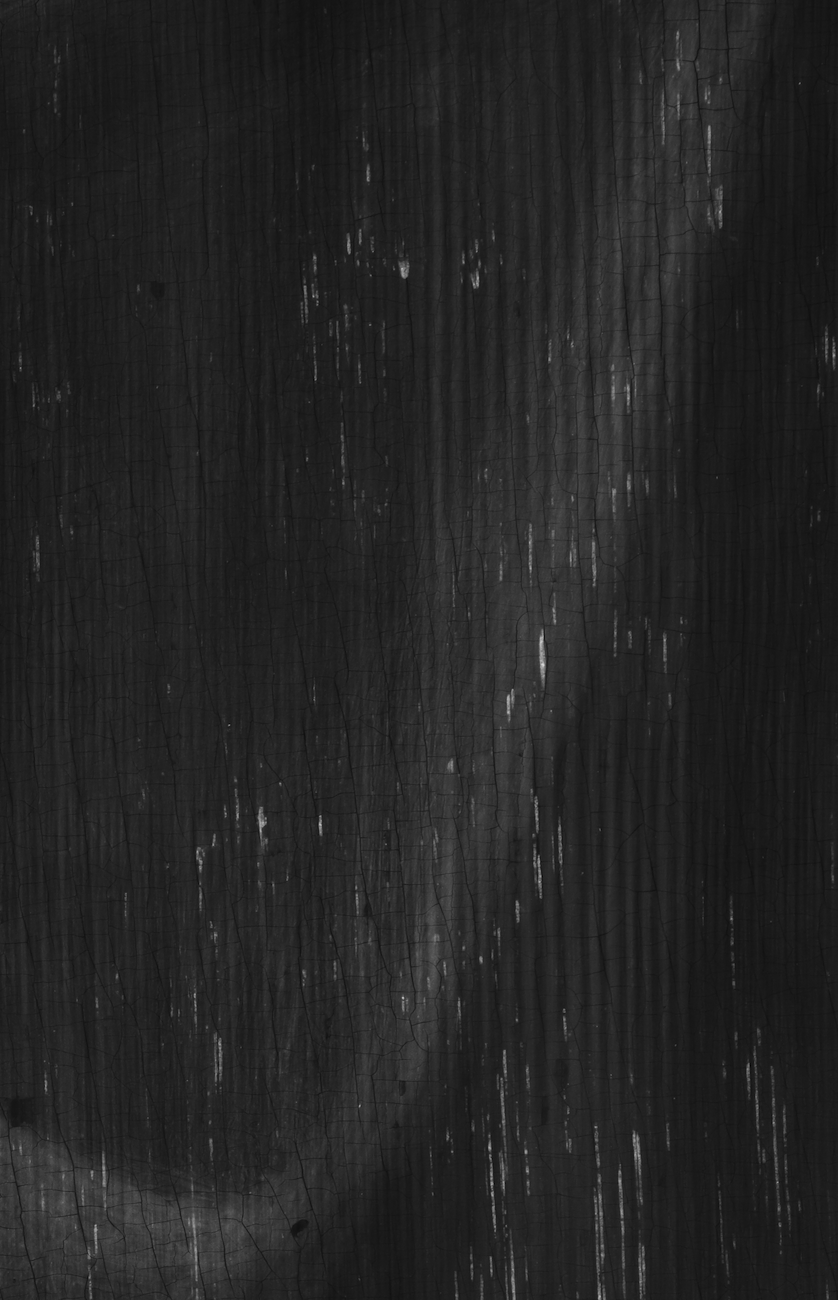}}
  \subfigure[Crack mask]{
    \label{fig:visualexample:c}
    \includegraphics[width=0.15\textwidth]{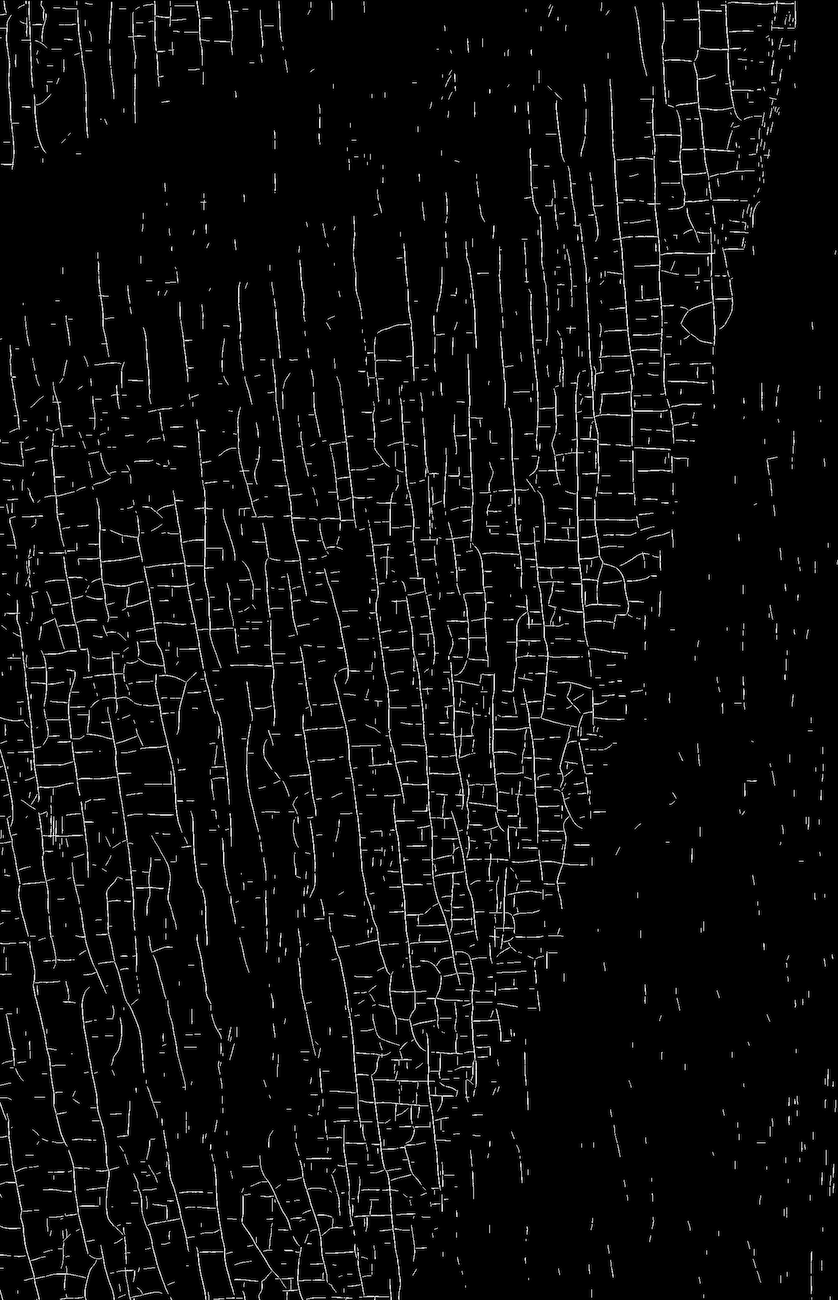}}
  \caption{Examples of images from single-sided panels of the \textit{Ghent
Altarpiece} and the corresponding crack masks.}
  \label{fig:visualexample}
\end{figure}

\begin{figure*}[t]
 \centering
 \subfigure[]{
    \label{fig:XrrSIvisuals:a}
    \includegraphics[width=0.321\textwidth]{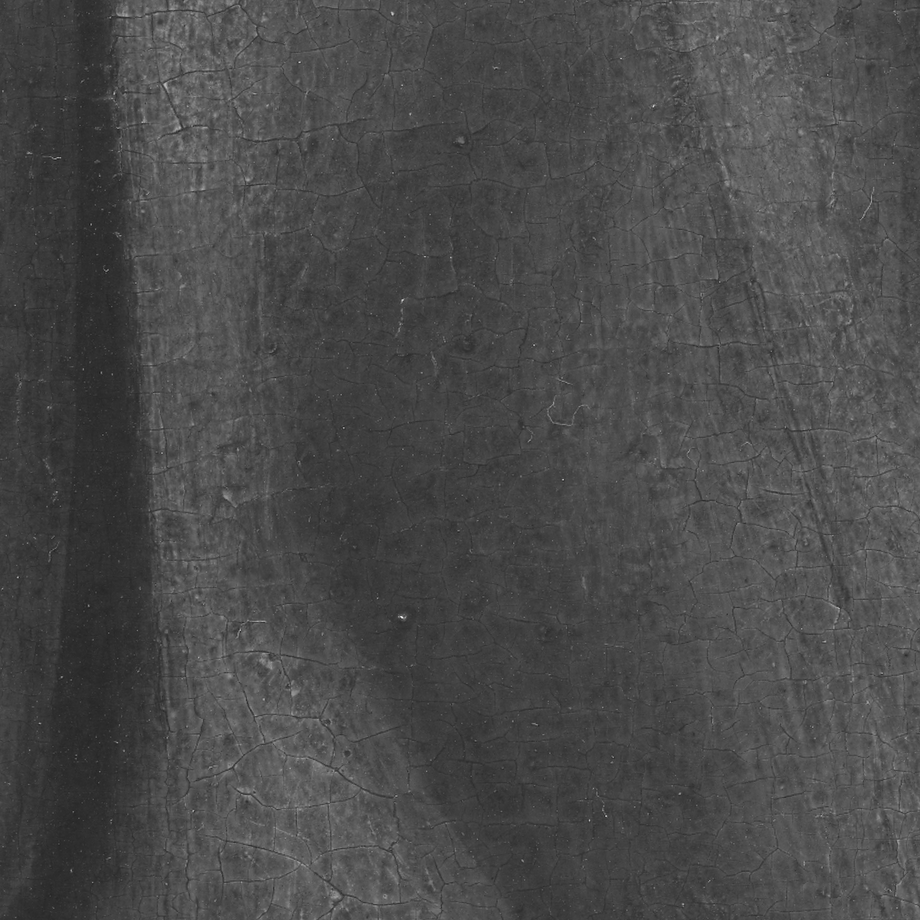}}
 \subfigure[]{
    \label{fig:XrrSIvisuals:b}
    \includegraphics[width=0.321\textwidth]{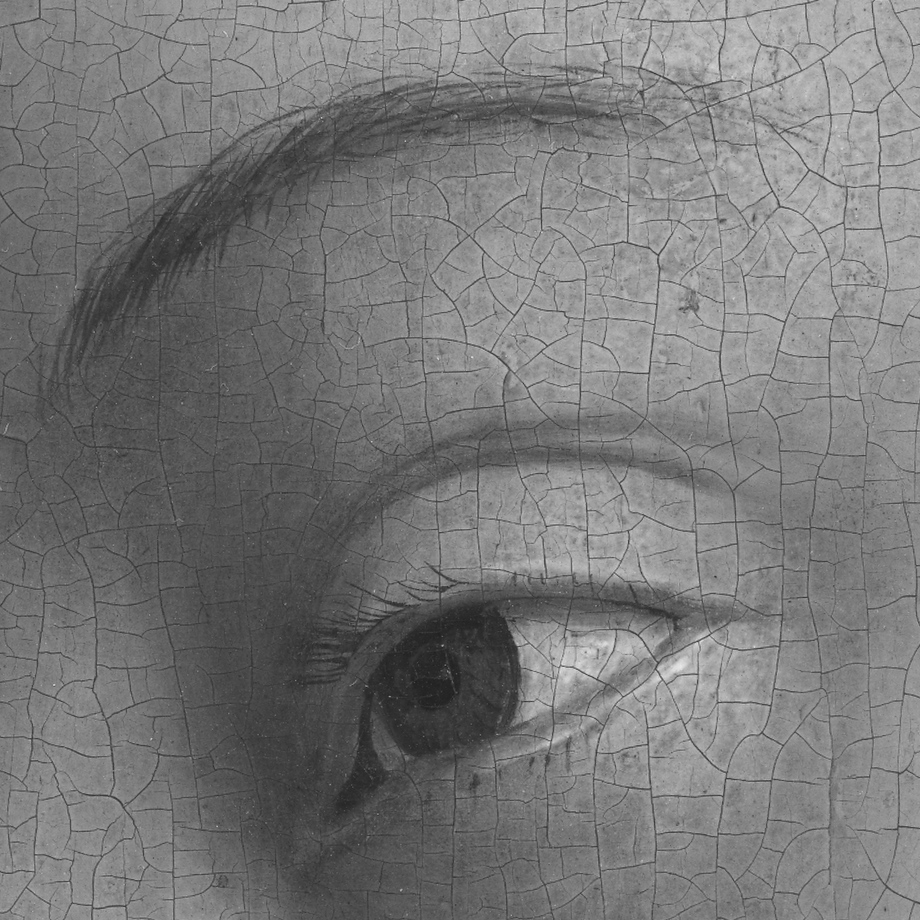}}
  \subfigure[\tiny\copyright KIK-IRPA] {
    \label{fig:XrrSIvisuals:a}
    \includegraphics[width=0.321\textwidth]{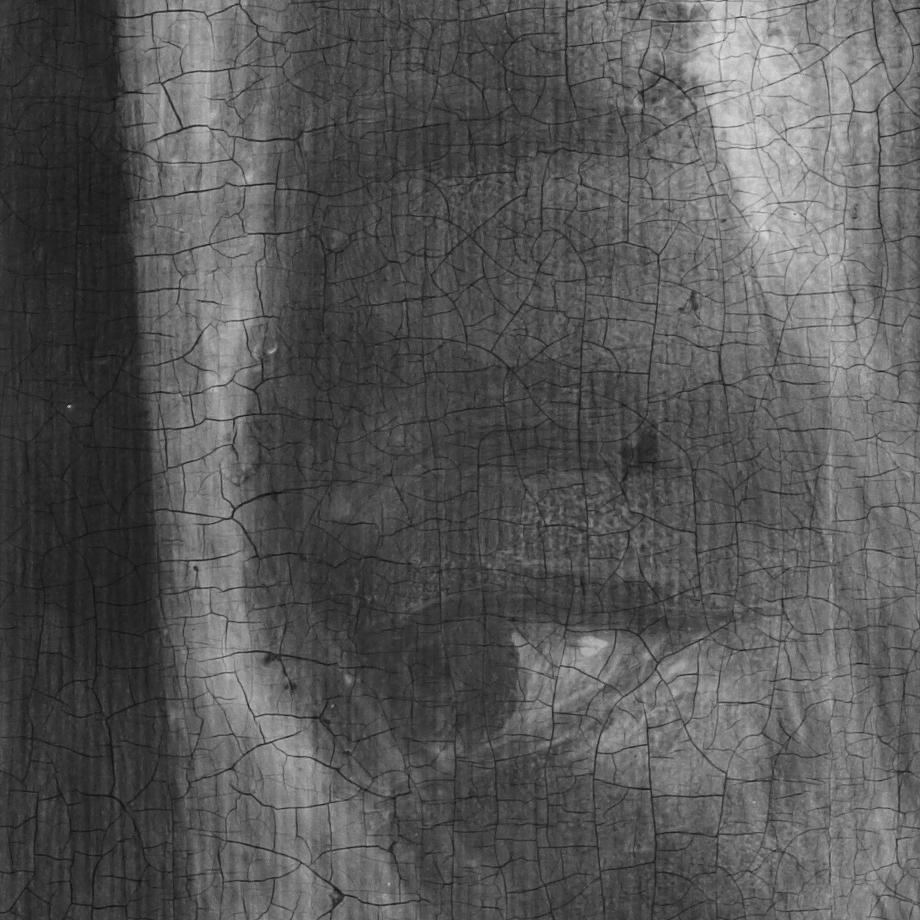}}
   \subfigure[]{
    \label{fig:XrrSIvisuals:a}
    \includegraphics[width=0.321\textwidth]{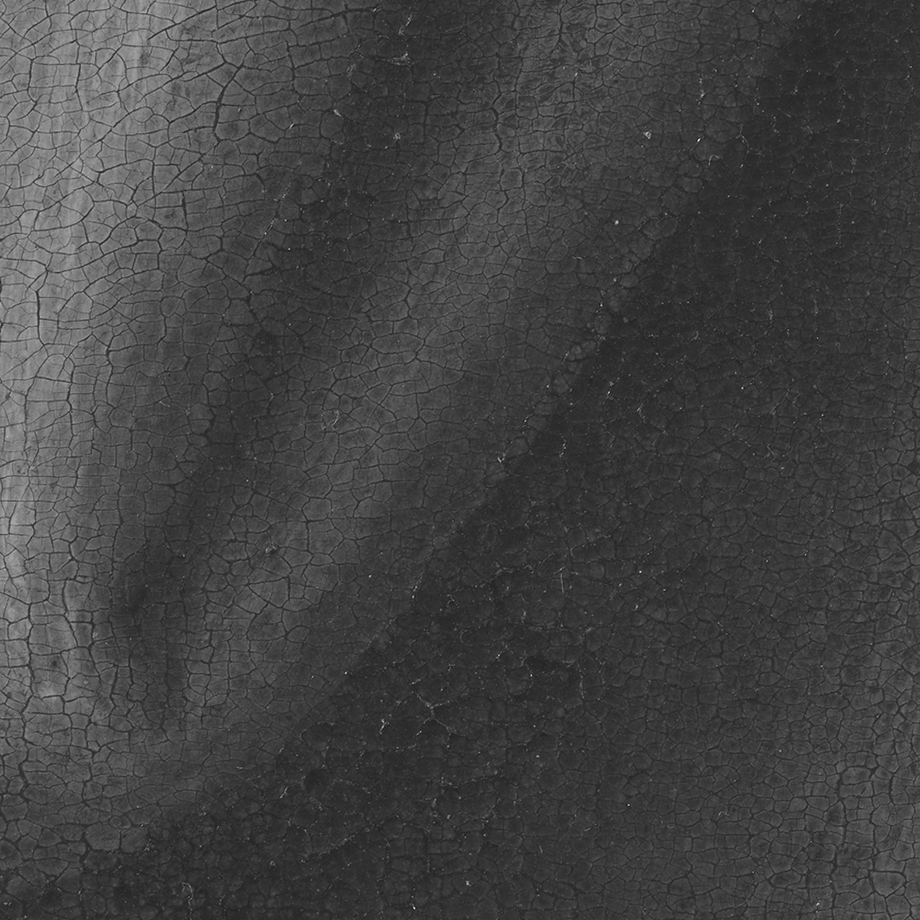}}
 \subfigure[]{
    \label{fig:XrrSIvisuals:b}
    \includegraphics[width=0.321\textwidth]{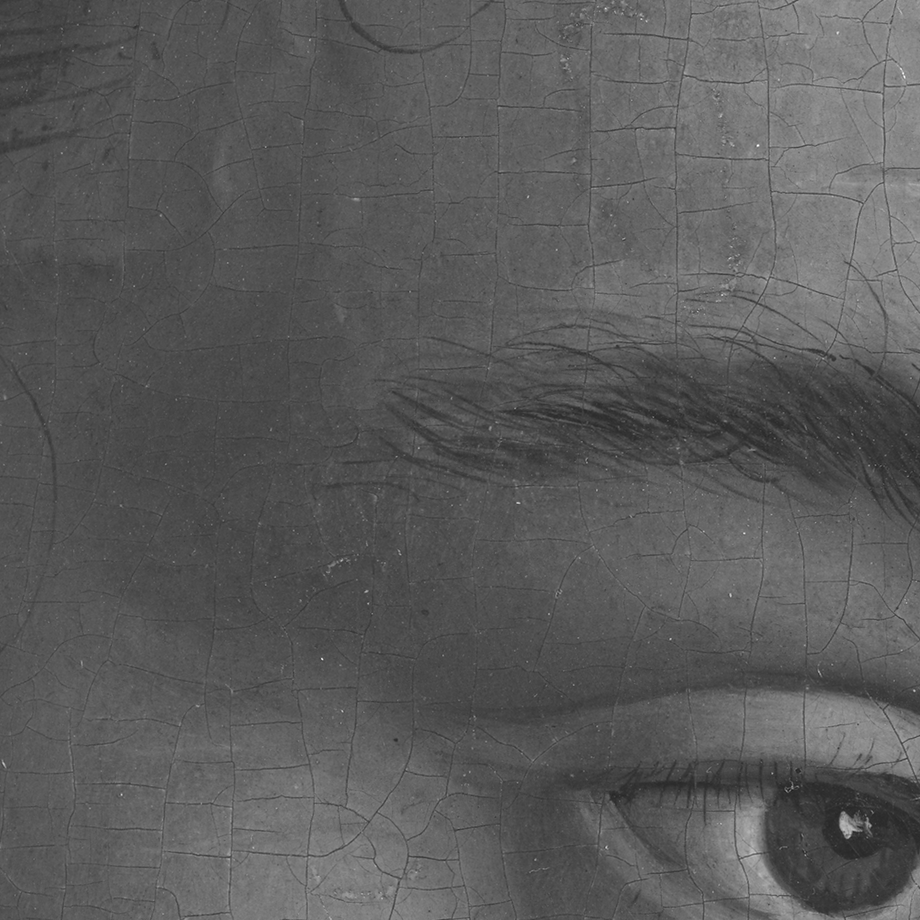}}
  \subfigure[\tiny\copyright KIK-IRPA] {
    \label{fig:XrrSIvisuals:a}
    \includegraphics[width=0.321\textwidth]{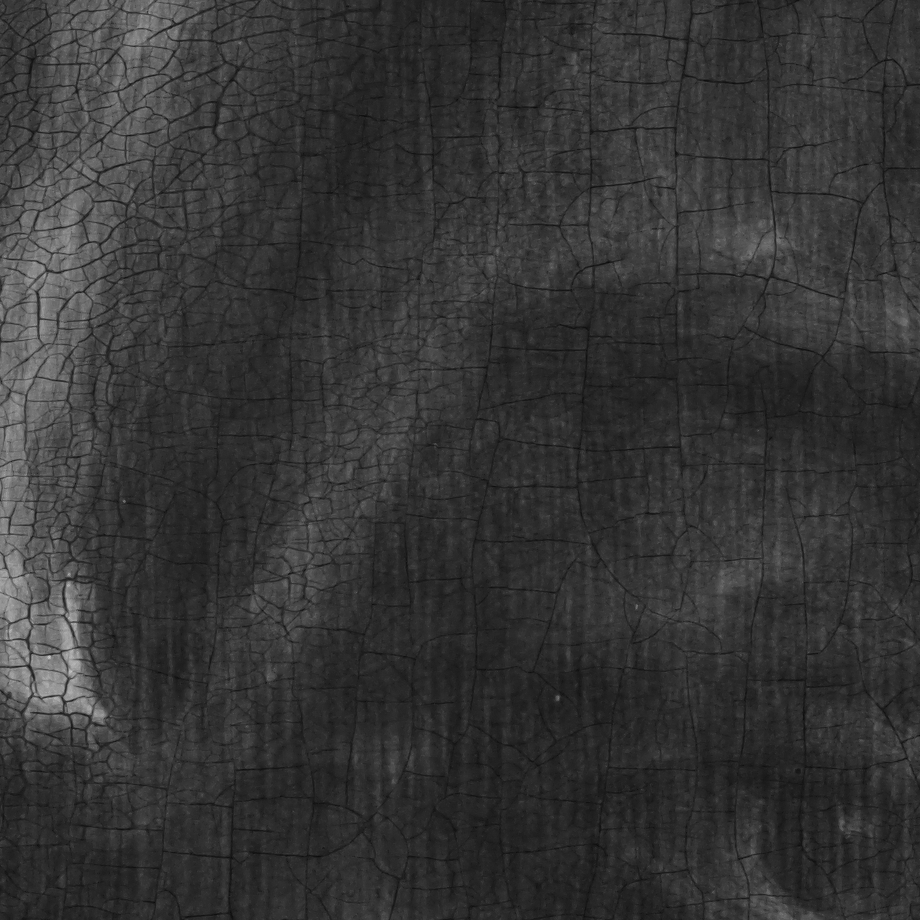}}
  \caption{Image set cropped from a double-sided panel of the altarpiece,
 on which we assess the proposed method; (a) and (d) photograph of side 1,
(b) and (e) photograph
of side 2; (c) and (f) corresponding  X-ray image. The resolution is $1024\times{1024}$
pixels.}
\label{fig:XrrSIvisuals}
\end{figure*}

To retrieve the initial dictionaries, we apply the proposed method in Section \ref{sec:coupledDicLearn} with the dictionaries  initialised
randomly and the maximum number of iterations  set to 100---experimental evidence has shown that this value strikes a good balance between complexity and dictionary identifiability. To compare
the retrieved dictionaries with the original ones, we adhere to the approach
in \cite{aharon2006img}: per generating dictionary, we sweep through its
columns and identify the closest column in the retrieved dictionary. The
distance between the two columns is measured as~$1-|\delta_i\tilde{\delta}^T_j|$,
where $\delta_i$ is the $i$-th column in the original dictionary $\Psi^c$,
$\Phi^c$, or $\Phi$, and $\tilde\delta_j$ is the $j$-th column in the corresponding
recovered dictionary. Similar to \cite{aharon2006img}, a distance less than
0.01  signifies a success. The percentage of successes per dictionary and for various SNR values is
reported in Table \ref{tab:DictionaryIdentifiability}. The results, which
are averaged over $100$ trials,  show that for very noisy data (that is, $\text{SNR}\leq15$)
the dictionary identifiability performance is  low. However, for  SNR values
higher than 20 dB, the percentage of recovered dictionary atoms is up to 
96.78\%. The obtained performance is systematic for different dictionary and
signal sizes as well as for different sparsity levels.  

In a second stage, given the learned dictionaries, we separate signal
pairs $(x_1,x_2)$ from mixtures $m=x_1+x_2$ by solving Problem \eqref{Eq:Prob}
using the corresponding pair $(y_1,y_2)$ as side information. The pairs are taken from the  correlated data signals $X$ and
$Y$, to which white Gaussian noise with a varying SNR
has been added. Table \ref{tab:SourceSeparation} reports the normalized mean-squared error between the reconstructed---defined by $\tilde{x_i}$---and the original signals, that is, $\frac{\|x_i-\tilde{x_i}\|_2^2}{\|x_i\|_2^2}$, $i=\{1,2\}$. The results show that at low and moderate noise SNRs the reconstruction error is very low. When the noise increases, the recovery performance drops; this is to be expected as the noise affects both the dictionary leaning and the generation of the mixtures.

\begin{figure*}[t]
\centering
\includegraphics[width=0.321\textwidth]{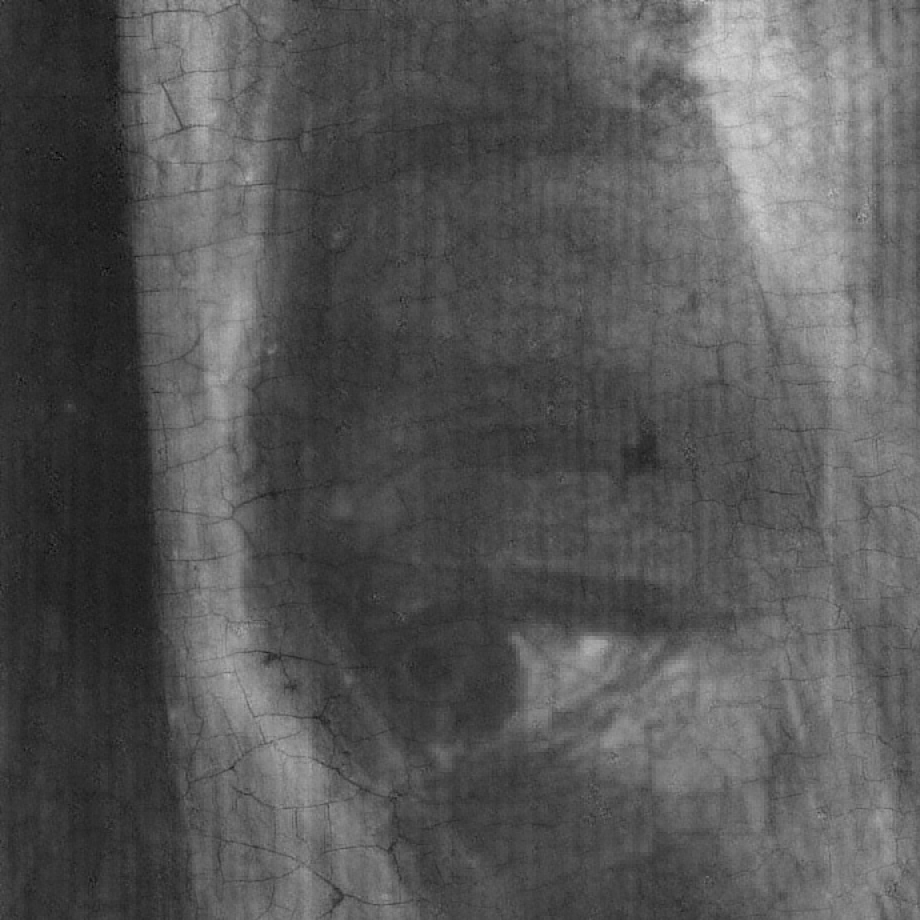}
\includegraphics[width=0.321\textwidth]{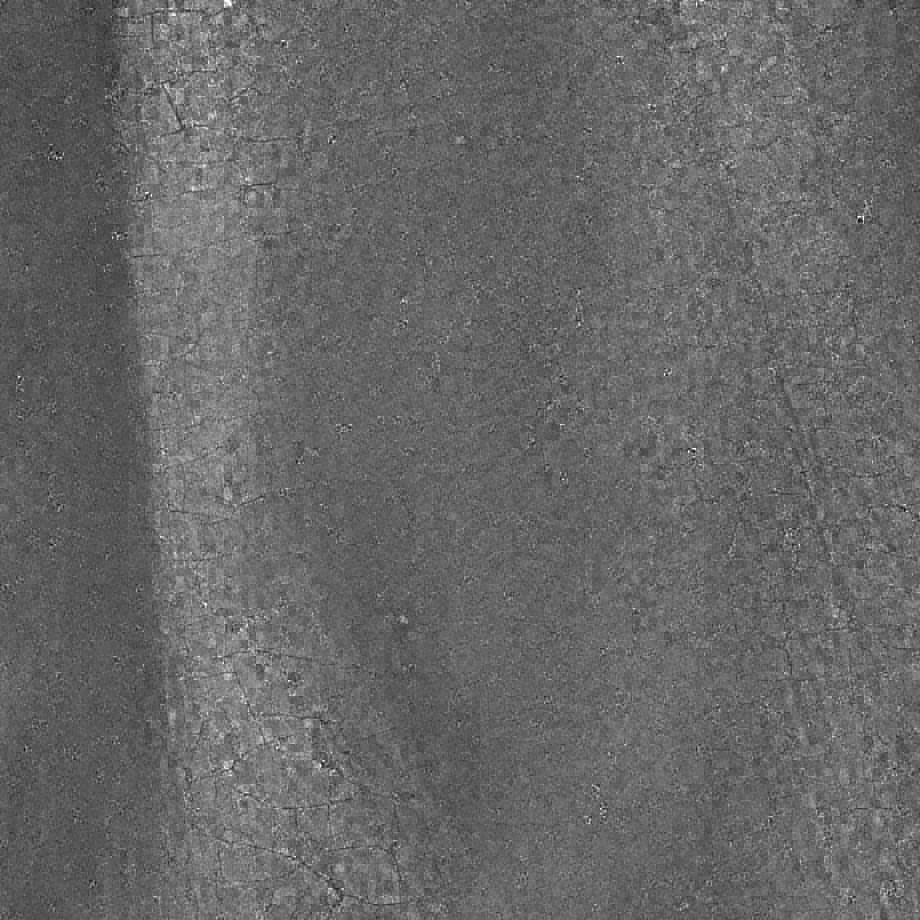}
\includegraphics[width=0.321\textwidth]{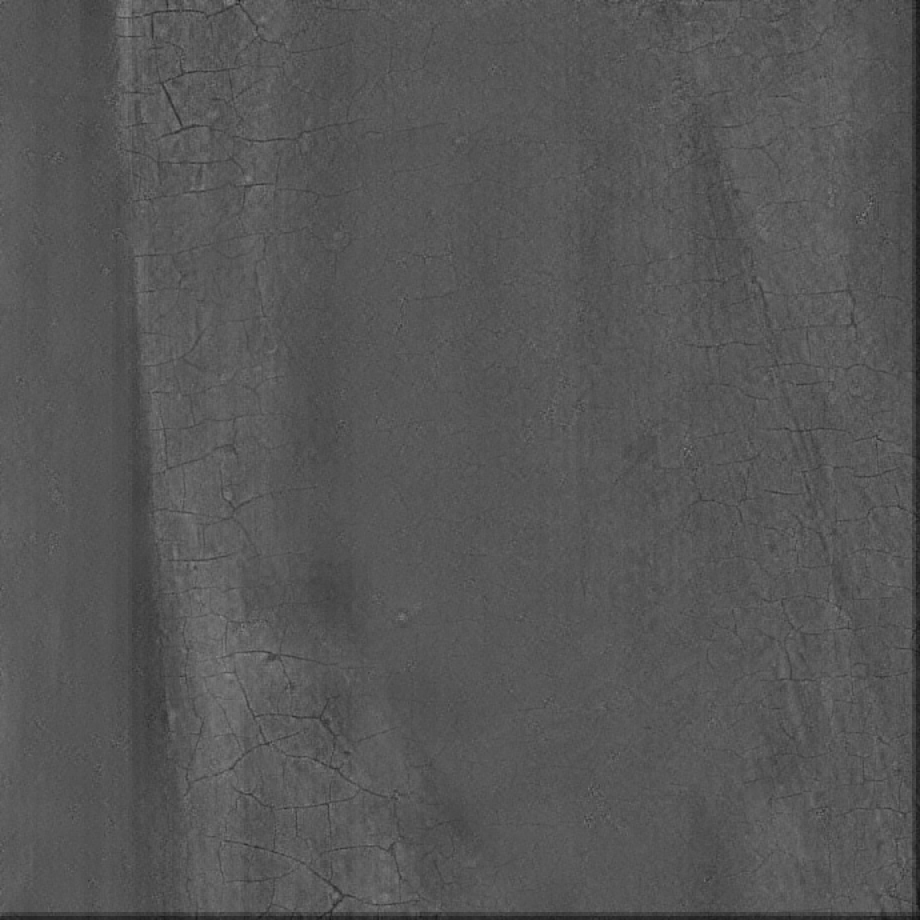}

\medskip

\includegraphics[width=0.321\textwidth]{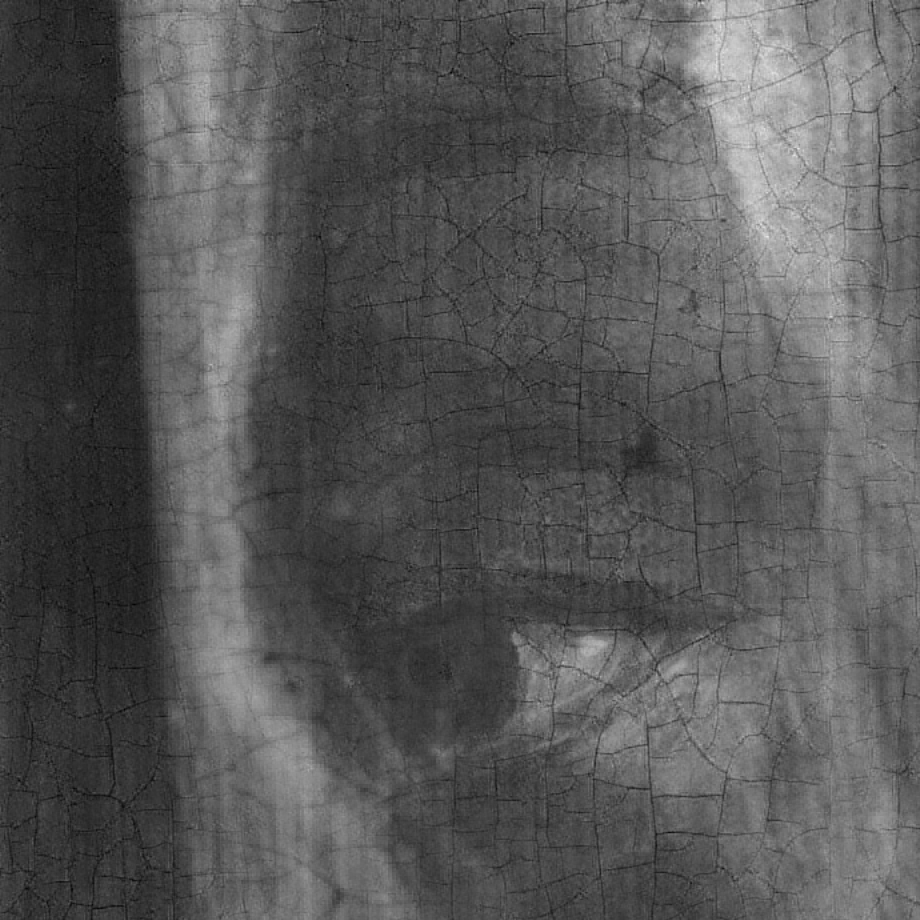}
\includegraphics[width=0.321\textwidth]{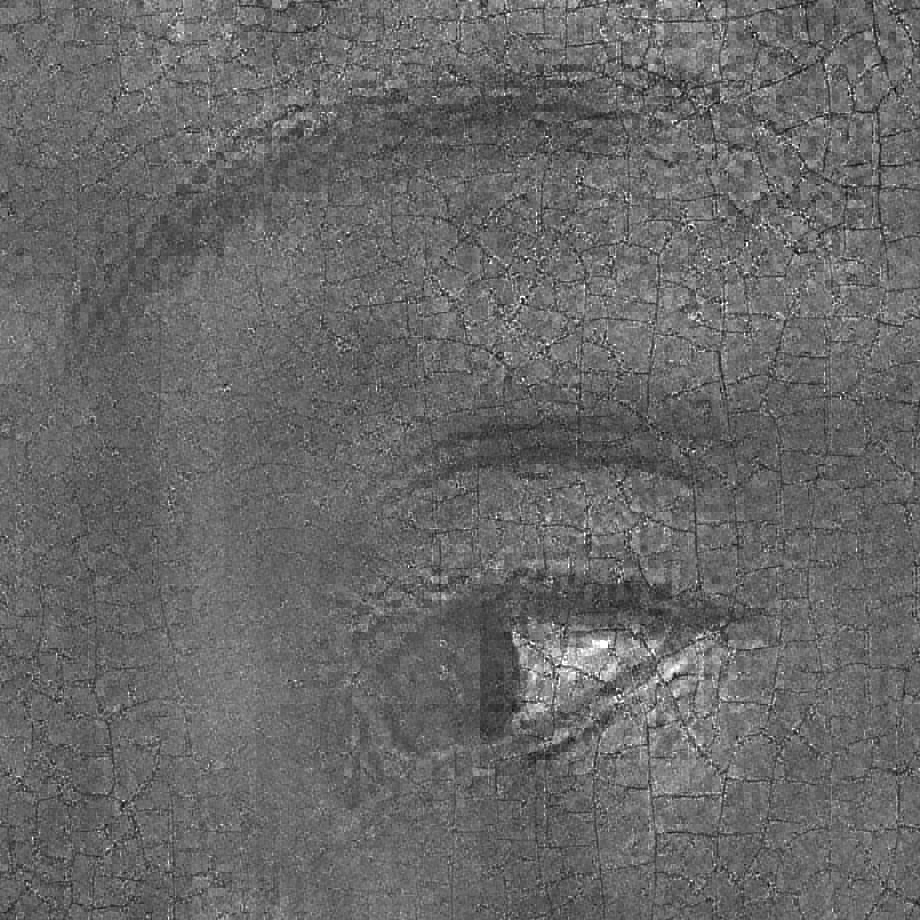}
\includegraphics[width=0.321\textwidth]{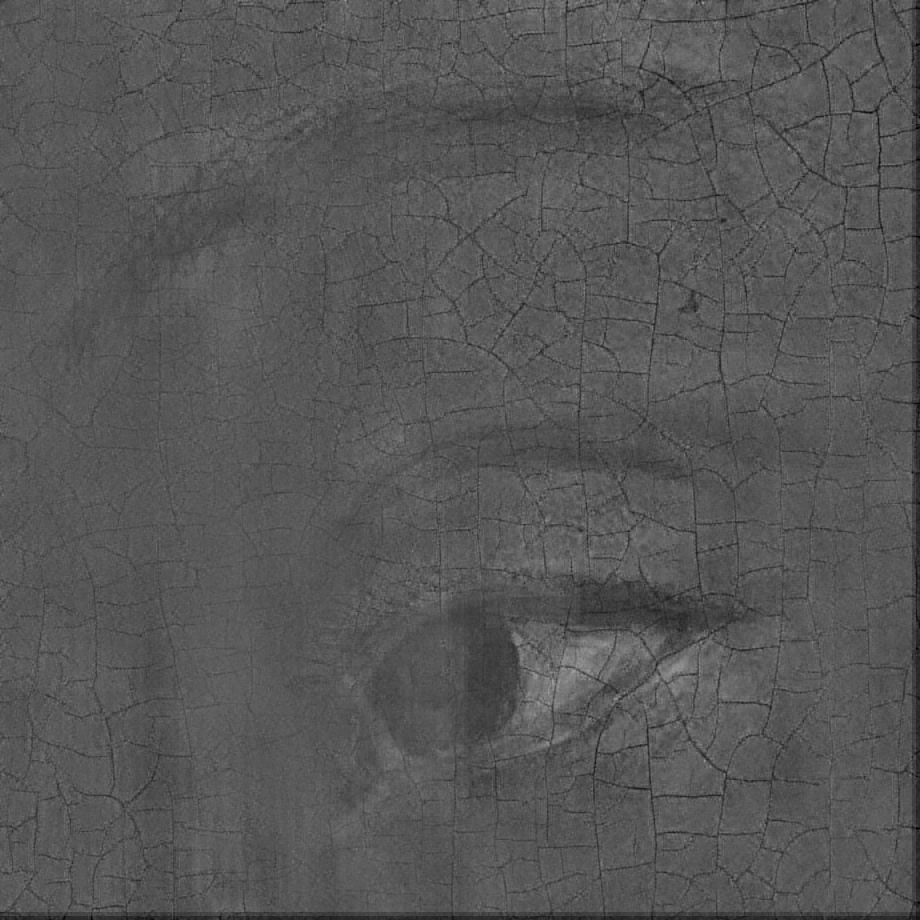}
\caption{Visual evaluation of the different configurations of the proposed
method in the separation of the X-ray image in Fig. \ref{fig:XrrSIvisuals}(c);
(first row) separated side 1, (second row) separated side 2. The configurations 
are: (first column) single-scale method (Section \ref{sec:SingScaleAppr}) with the
coupled dictionary learning algorithm described in Section~\ref{sec:coupledDicLearn},
(second column) multi-scale method (Section~\ref{sec:MultiScaleSubSection}) with
the
 coupled dictionary learning method from Section~\ref{sec:coupledDicLearn},
(third column) multi-scale method (Section~\ref{sec:MultiScaleSubSection}) with
the
 weighted coupled dictionary learning method from Section~\ref{sec:weightedcoupledDicLearn} and without including the $v$ component.}
\label{fig:visualOurImprovements}
\end{figure*}

\begin{figure}[t]
 \centering
  \subfigure[]{
    \label{fig:visualComparisonSOTA1:a}
    \includegraphics[width=0.323\textwidth]{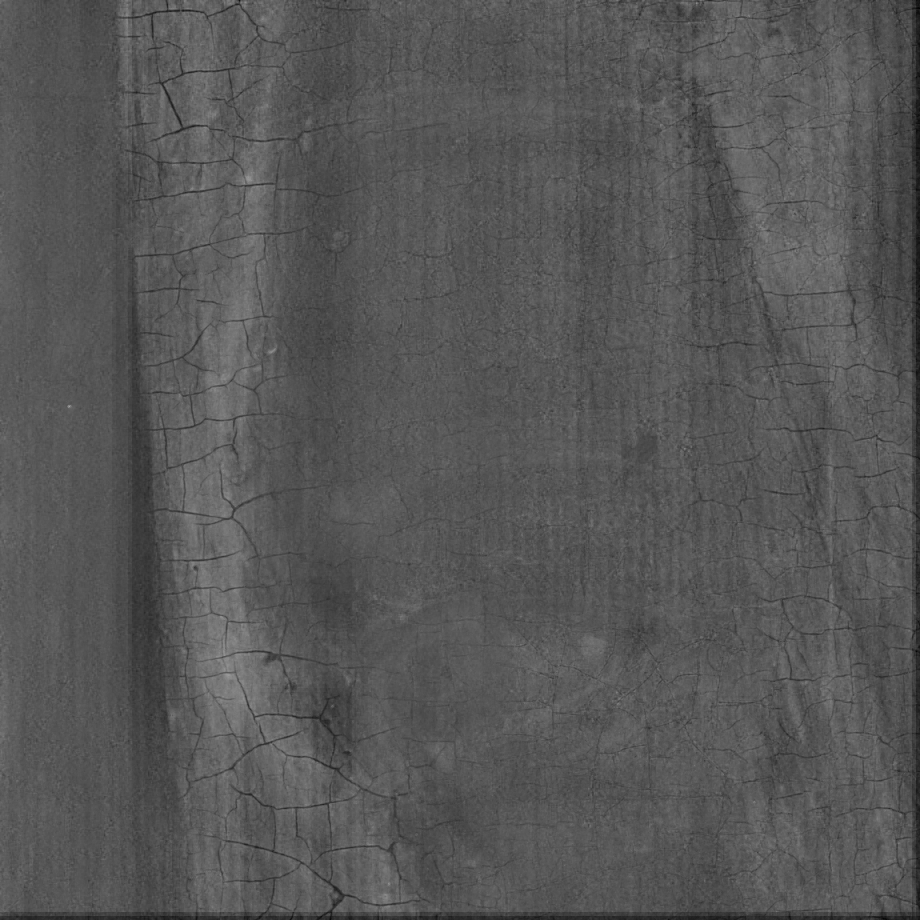}}
  \subfigure[]{
    \label{fig:visualComparisonSOTA1:b}
    \includegraphics[width=0.323\textwidth]{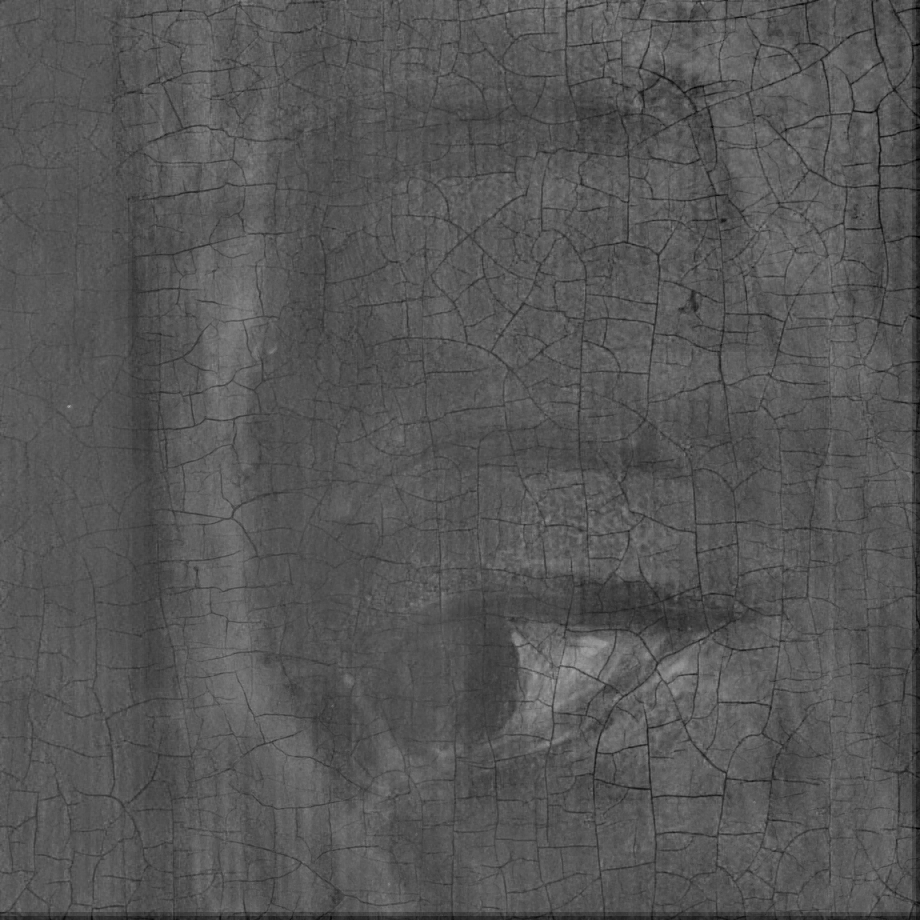}}
\caption{Visual evaluation of the proposed multi-scale
method in the separation of the X-ray image  in Fig. \ref{fig:XrrSIvisuals}(c);
(a) separated side 1, (b) separated side 2. The reconstructions include the
X-ray specific $v$ component.}
\label{fig:visualComparisonWithVComponent}
\end{figure}

\begin{figure*}[t]
\centering
\includegraphics[width=0.321\textwidth]{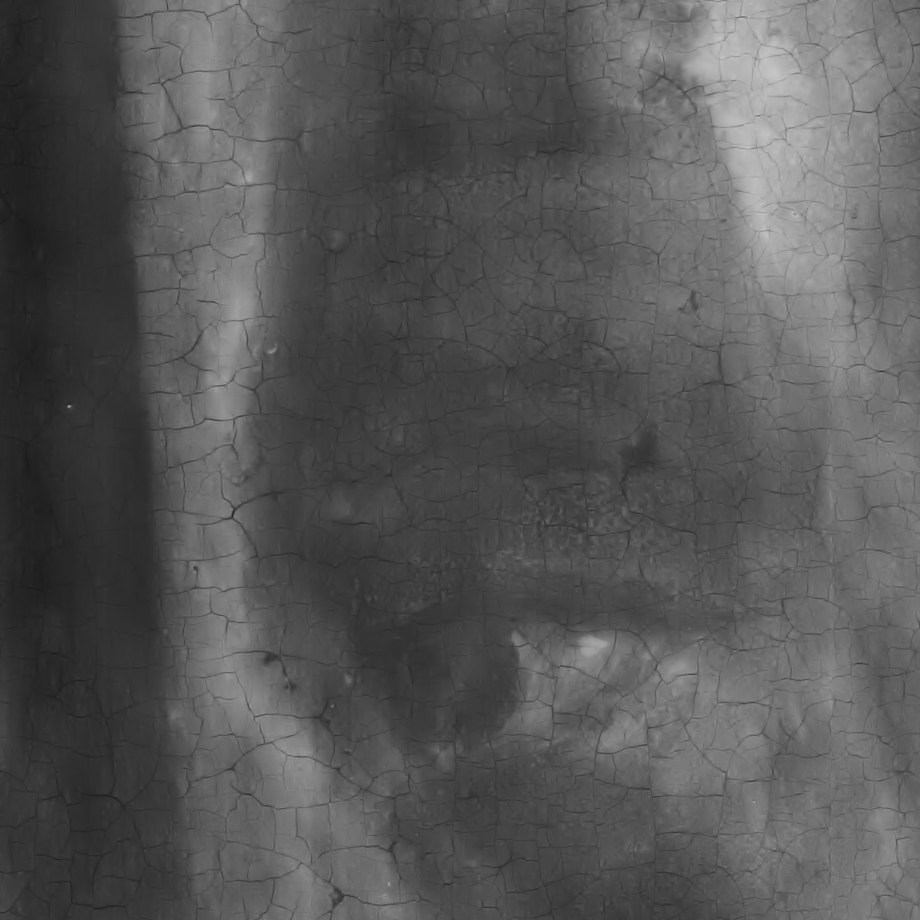}
\includegraphics[width=0.321\textwidth]{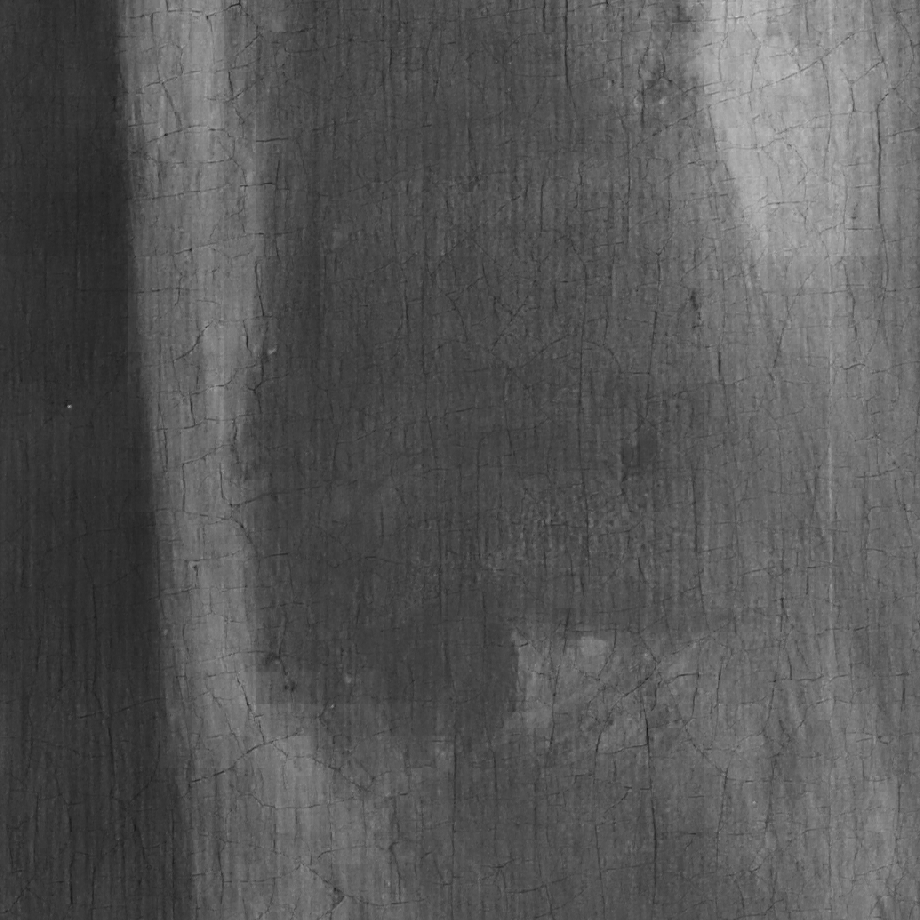}
\includegraphics[width=0.321\textwidth]{sep1noV_exp1_flt8}

\medskip
\includegraphics[width=0.321\textwidth]{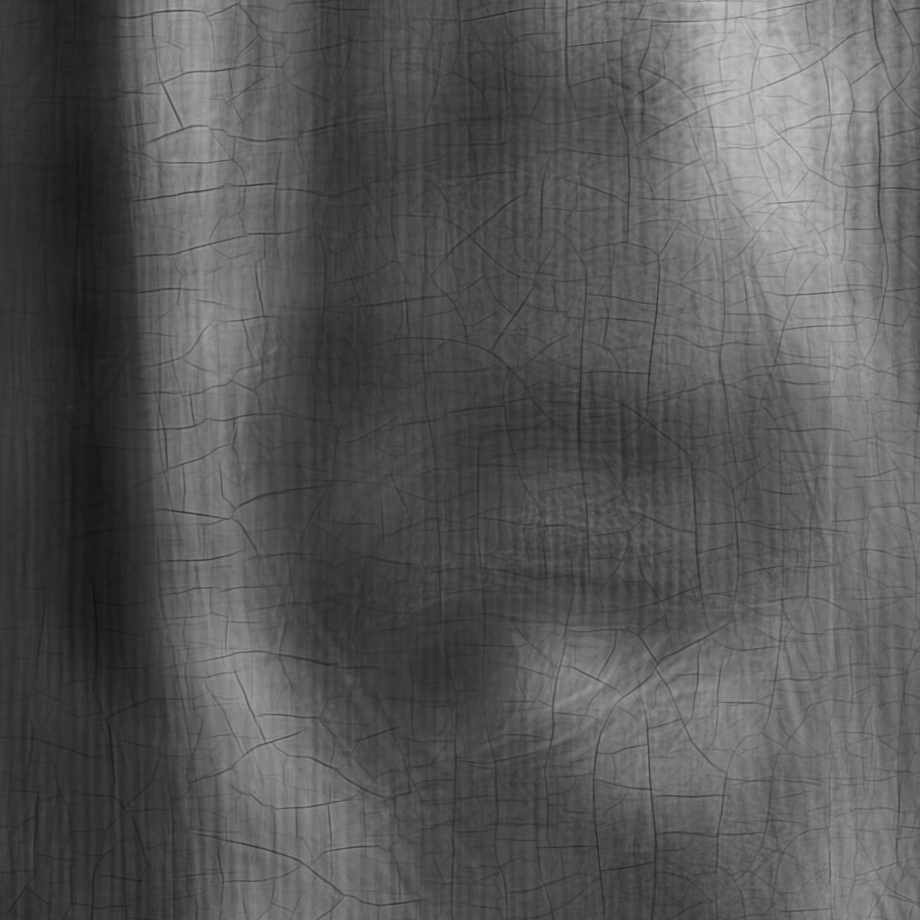}
\includegraphics[width=0.321\textwidth]{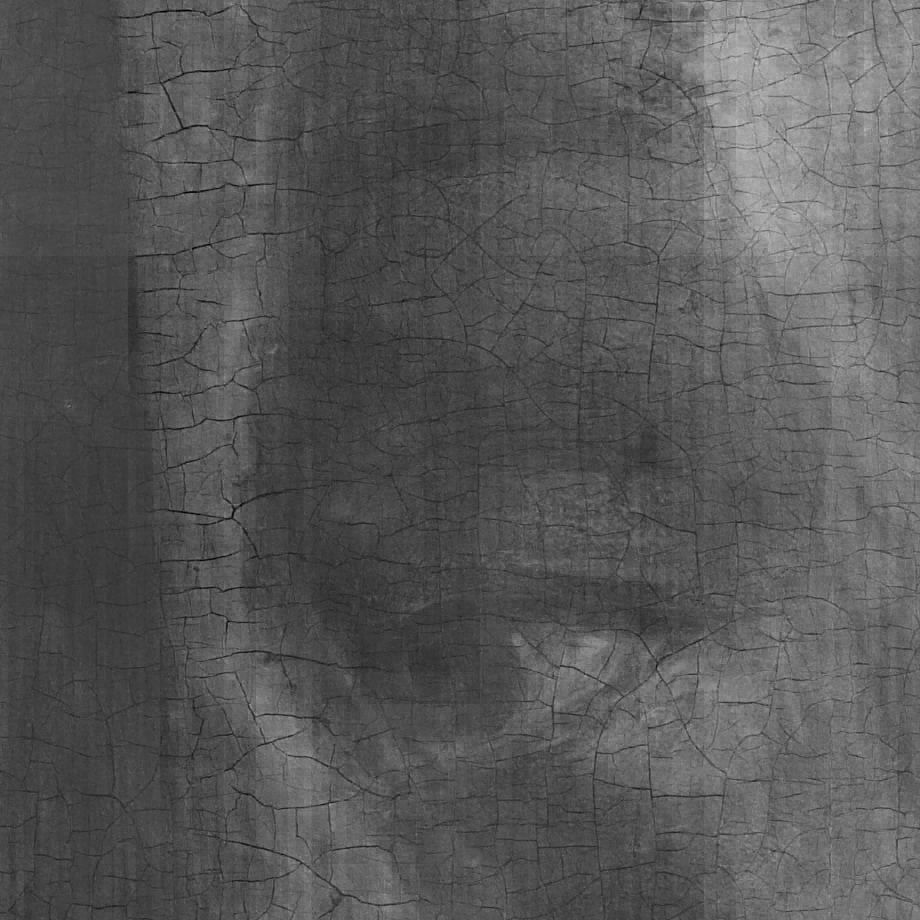}
\includegraphics[width=0.321\textwidth]{sep2noV_exp1_flt8}
  \caption{Visual evaluation of the proposed multi-scale
method in the separation of the X-ray image in Fig. \ref{fig:XrrSIvisuals}(c);
(first row) separated side 1, (second row) separated side 2. The competing methods
are: (first column) MCA\ with fixed dictionaries \cite{icip14}, (second column) multi-scale
MCA with K-SVD, (third column) Proposed without including the $v$ component.}
\label{fig:visualComparisonSOTA1}
\end{figure*}


\begin{figure*}[t]
\centering
\includegraphics[width=0.321\textwidth]{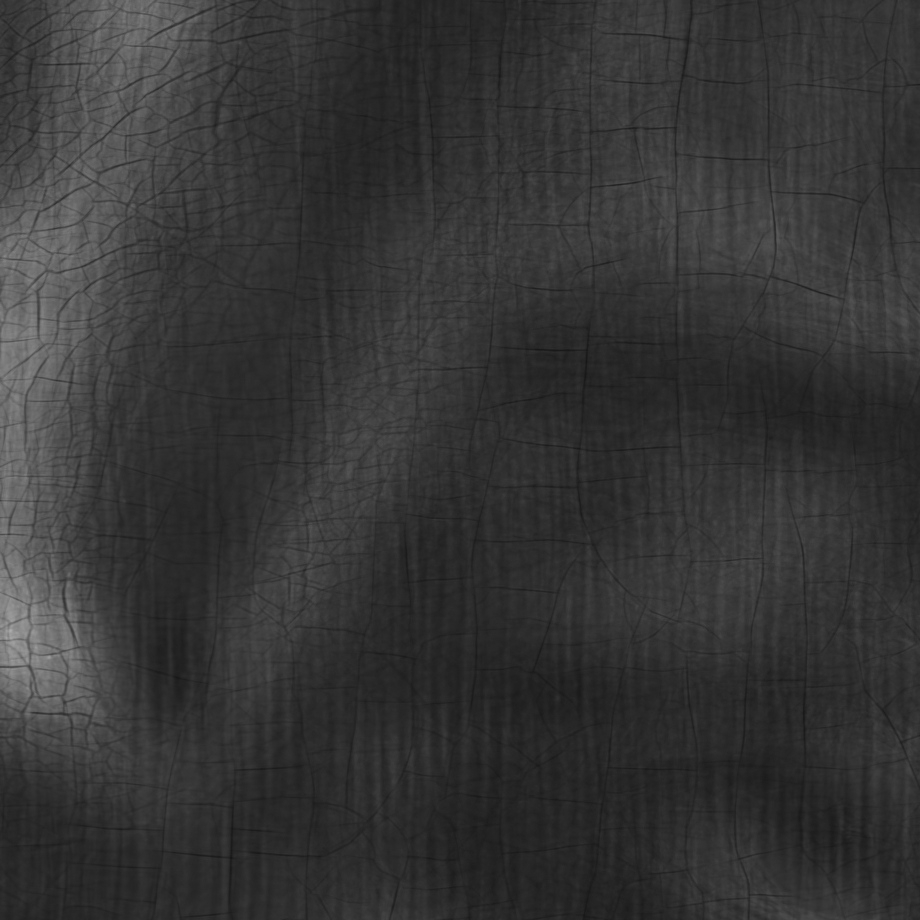}
\includegraphics[width=0.321\textwidth]{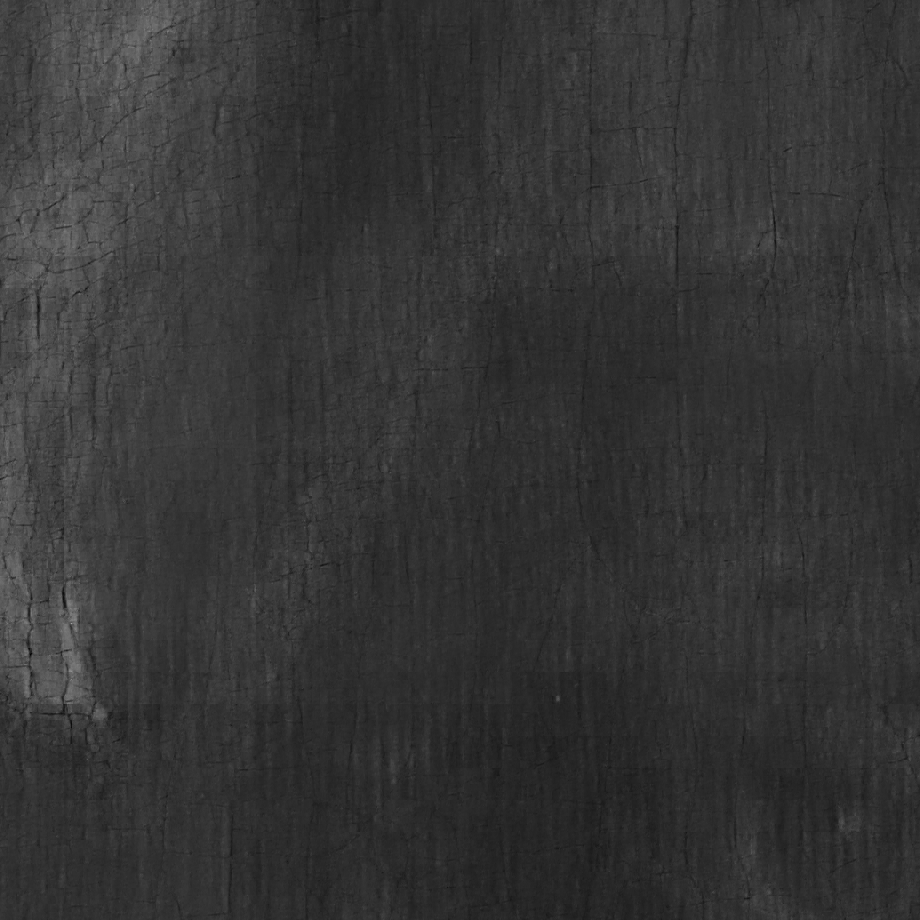}
\includegraphics[width=0.321\textwidth]{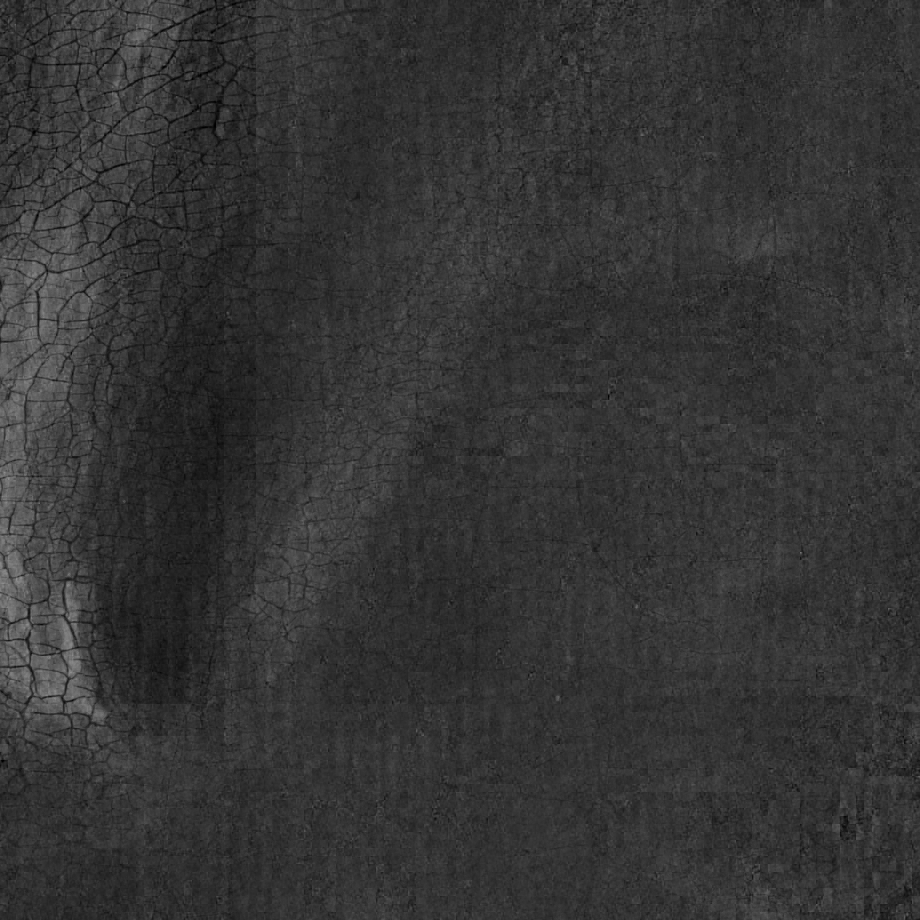}

\medskip
\includegraphics[width=0.321\textwidth]{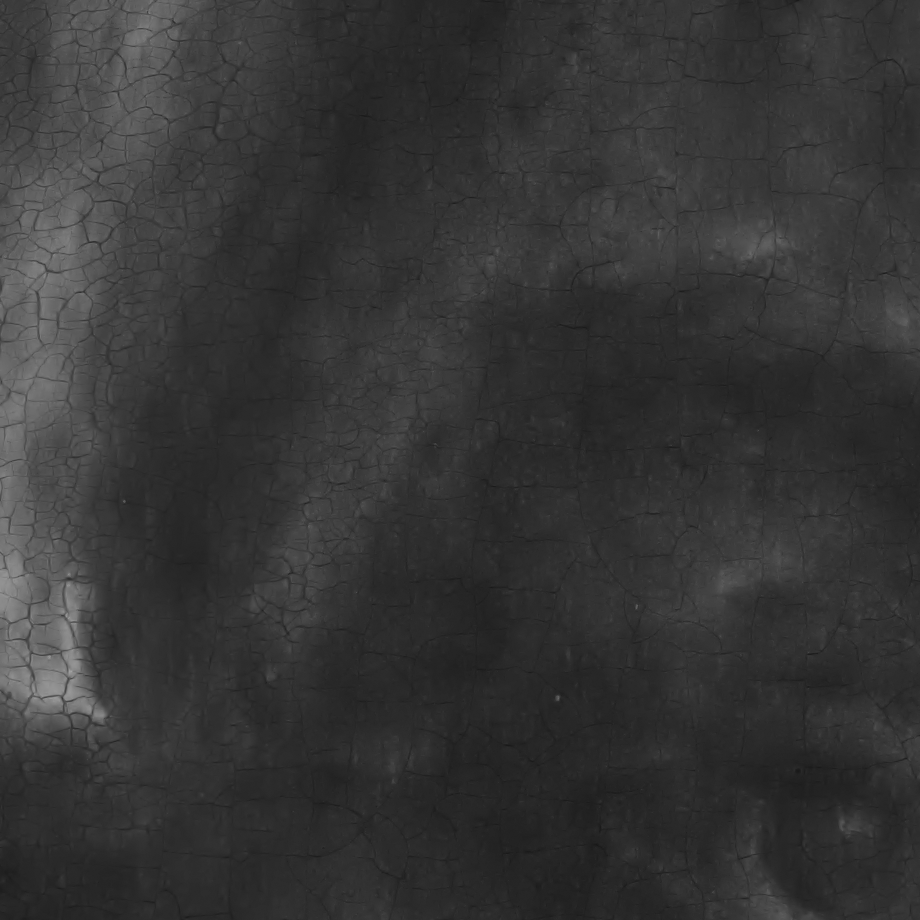}
\includegraphics[width=0.321\textwidth]{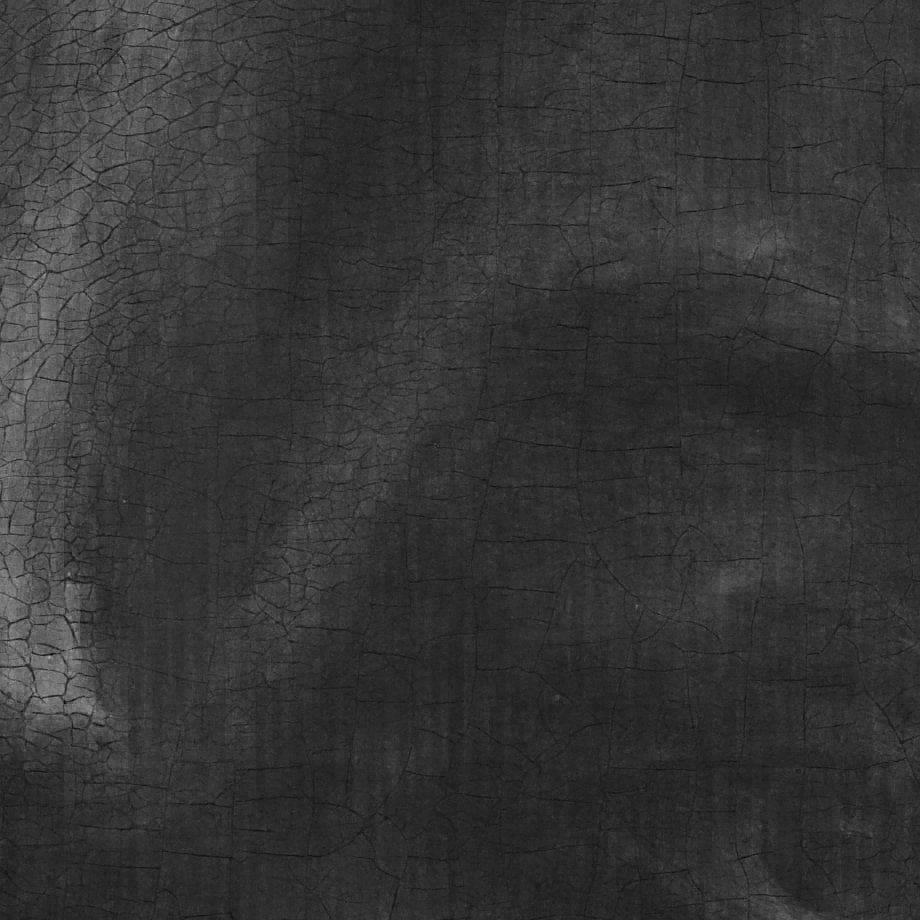}
\includegraphics[width=0.321\textwidth]{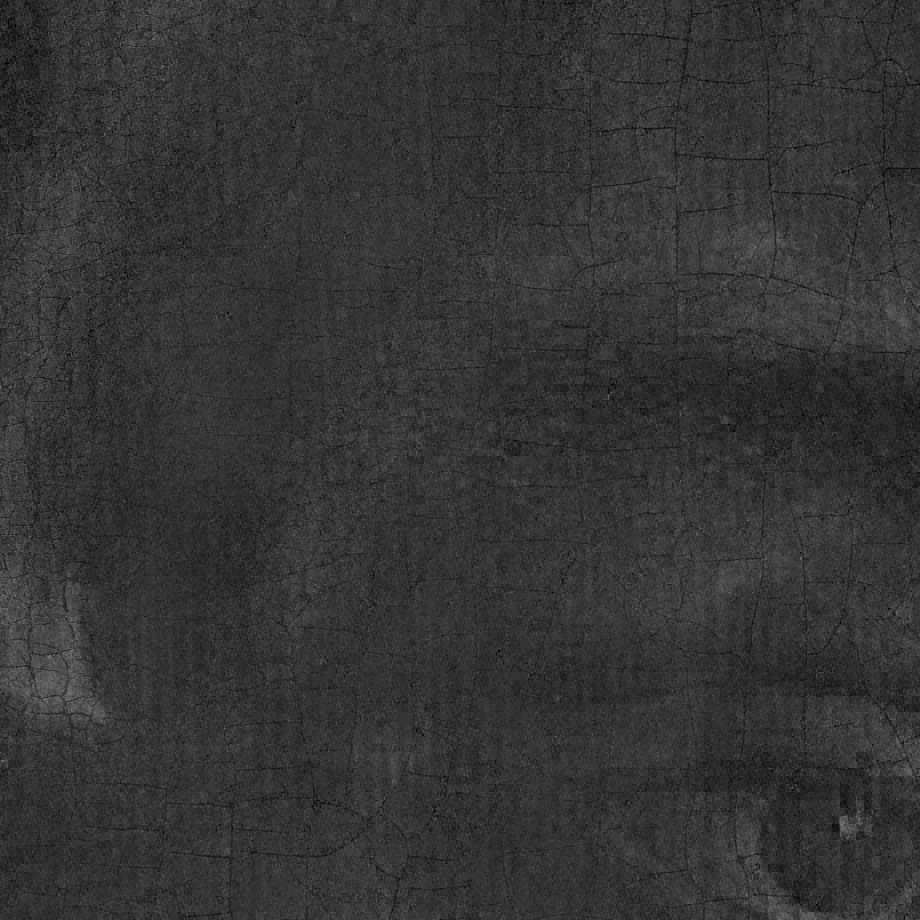}
  \caption{Visual evaluation of the proposed multi-scale
method in the separation of the X-ray image in Fig. \ref{fig:XrrSIvisuals}(f);
(first row) separated side 1, (second row) separated side 2. The competing
methods
are: (first column) MCA\ with fixed dictionaries \cite{icip14}, (second column)
multi-scale
MCA with K-SVD, (third column) Proposed including the $v$ component.}
\label{fig:visualComparisonSOTA2}
\end{figure*}


\subsection{Experiments with Real Data}
\label{sec:ExpsWithData} 
We consider eight image pairs---each consisting of an X-ray scan and the corresponding photograph---taken from digital acquisitions~\cite{pizurica2015digital} of single-sided panels of the \textit{Ghent Altarpiece} (1432). Furthermore, we are given access to eight crack masks (one per visual/X-ray image pair) that indicate the pixel positions referring to cracks (these masks were obtained using our method in \cite{cornelis2012crack}). Fig. \ref{fig:visualexample} depicts two such pairs with the crack masks, one visualizing a face and the other a piece of fabric. An example X-ray mixture (of size $1024\times1024$ pixels) together with its two registered visual images  corresponding to the two sides of the painting are depicted in Fig. \ref{fig:XrrSIvisuals}.

Firstly, adhering to the single-scale approach, described in Section \ref{sec:SingScaleAppr}, we train a dictionary triplet, $(\Psi^c,\Phi^c,\Phi) $, using our method in Section \ref{sec:coupledDicLearn}. We use $t=46400$ patches, each containing  $8\times{8}$ pixels, the  dictionaries, $\Psi^c$, $\Phi^c$, $\Phi$, have a dimension of $64\times{256}$, and we  set $s_{z}=10$ and $s_{v}=8$. The separated X-rays that correspond to the mixture in Fig. \ref{fig:XrrSIvisuals} are depicted in the first column of Fig. \ref{fig:visualOurImprovements}. We observe that our single-scale approach separates the texture of the X-rays; this is demonstrated by the accurate separation of the cracks. Still, however, the low-pass band content is not properly split over the images; namely, part of the cloth and the face are present in both separated images.  Next, we apply the  multi-scale framework, where we use~$L=4$ scales with parameters  $\sqrt{n_l} = 8, \: l=\{1, 2, 3, 4\}$, $\epsilon_1 = 4$, $\epsilon_2=4$,  $\epsilon_3= 7$, and $\epsilon_4= 8$. Dictionary triplets $(\Psi^c_\ell,\Phi^c_\ell,\Phi_\ell)$, each with dimension of $64\times{256}$, are trained for the first three scales and the dictionaries of the third scale are used for the forth. We use $t_1=46400$, $t_2=46400$ and $t_3=35500$ patches for scale 1, 2 and 3, respectively. The visualizations in the second column of Fig.~\ref{fig:visualOurImprovements} show that, compared to the single scale approach, the multi-scale method properly discriminates the low-pass frequency content of the two images (most part of the cloth is allocated to \textquotedblleft Separated Side 1\textquotedblright\ while the face is only visible in \textquotedblleft Separated Side 2\textquotedblright), thereby leading to a higher separation performance. Finally, we also
construct dictionary triplets according to our weighted dictionary learning
method in Section \ref{sec:weightedcoupledDicLearn}. The remaining dictionary
learning parameters are as before.   It is worth mentioning that, in order
to obtain a  solution in \eqref{eq:updateDicCol}, the number of training
samples $t$ needs to be higher that the total dimension of the dictionary.
Namely, to update the columns of dictionary $\Psi^c$ we need at least $16384$
samples. Correspondingly, to update the rows of dictionary $\overline{\Phi}$
we need more than $32768$ samples. The visual results in the third column of Fig.~\ref{fig:visualOurImprovements} corroborate
that the  quality of the separation is improved when the dictionaries are learned from only non-crack pixels. Indeed, with this configuration, the separated images
are not only smoother but also the separation is more prominent.

It is worth mentioning that the results of our method, depicted in Fig. \ref{fig:visualOurImprovements}, are obtained without including the  $v$ component during the reconstruction; namely,
we  reconstructed each  X-ray patch as $x_1 = \Phi^{c}z_{1c}$ and $x_2= \Phi^{c}z_{2c}$.
The visual results of our method when including the  $v$ component during the reconstruction are depicted in Fig.~\ref{fig:visualComparisonWithVComponent}. These results are obtained with the same dictionaries that yield the result in  the third column of Fig.
\ref{fig:visualOurImprovements}. By comparing the two reconstructions, we can make the following observations. First, the $v$ component  successfully expresses the  X-ray specific features, such as the wood grain, visualized by the periodic vertical stripes in the X-ray scan.  The reconstruction of these stripes is much more evident in Fig.~\ref{fig:visualComparisonWithVComponent}. Secondly, in this case, the $v$ component  also captures parts of the actual content that we wish to separate. For example, we can discern a faint outline of the eye in Fig.~\ref{fig:visualComparisonWithVComponent}(a) as well as a fold of fabric appearing in Fig.~\ref{fig:visualComparisonWithVComponent}(b).

We compare our best performing multi-scale approach (namely, the one that omits cracks when learning dictionaries) with the state-of-the-art MCA\ method \cite{starck2004redundant,bobin2007sparsity}. Two configurations of the latter are considered. Based on prior work~\cite{icip14}, in  one configuration we use fixed dictionaries, namely, the discrete wavelet and curvelet transforms are applied on blocks of $512\times 512$ pixels. Inherently, the low-frequency content cannot be split by MCA and it is equally divided between both retrieved components. In the other configuration, we learn dictionaries with K-SVD using the same training X-ray images as in the previous experiment.   One dictionary is trained on the X-ray images depicting  fabric and the other on the images of faces. The K-SVD\ parameters are the same as the ones used in our method. Furthermore, the same multi-scale strategy is applied to the configuration
of  MCA with  K-SVD trained dictionaries. The results are depicted in Fig. \ref{fig:visualComparisonSOTA1} and Fig. \ref{fig:visualComparisonSOTA2}. Note that the third column in  Fig.
\ref{fig:visualComparisonSOTA1} and Fig. \ref{fig:visualComparisonSOTA2} are without and with taking the $v$ component into account, respectively. It is clear that MCA with fixed dictionaries can only separate based on morphological properties; for example, the wood grain of the panel is captured  entirely by curvelets and not by the wavelets. It is, however, unsuitable to separate painted content---it is evident that part of the cloth and face appear in both separated components. Furthermore, MCA with K-SVD  dictionaries is also unable to separate the X-ray content. Nevertheless, we do observe that most cracks are captured by the face dictionary, as  more cracks are present in that type of content. Unlike both state-of-the-art configurations of MCA, the proposed method separates the X-ray content accurately (the cloth is always depicted on \textquotedblleft Separated Side 1\textquotedblright\ while the face is
only visible in \textquotedblleft Separated Side 2\textquotedblright), leading  to better visual performance. These results corroborate the benefit of  using side information by means of photographs to separate mixtures of X-ray images.

\subsection{Experiments on Simulated Mixtures}
Due to the lack of proper ground truth data, we generate simulated X-ray image mixtures in an attempt to assess our method in an objective manner. To this end, we utilised the X-ray images from single-sided panels, depicting content similar to the mixture in Fig. \ref{fig:XrrSIvisuals}(c) and (f). We generated mixtures by summing these independent X-ray images\footnote{We divided the sum by two to bring the mixture to the same range as the independent components.} and then we assessed the separation performance of the proposed method vis-\`{a}-vis  MCA\  either with fixed or K-SVD\ trained dictionaries. For this set of experiments, patches of size $256\times256$ pixels were considered and the parameters of the different methods were kept the same as in the previous section. Table~\ref{tab:SimulMixtureTable} reports the quality of the reconstructed X-ray components by means of the peak-signal-to-noise-ratio (PSNR) and structural similarity index metric (SSIM) \cite{wang2004image}. It is clear that the proposed method  outperforms the alternative state-of-the-art methods both in terms of  PSNR\ and  SSIM\ performance. Compared to MCA\ with fixed dictionaries, the proposed method brings an  improvement in the quality of the separation by up to $1.26$dB in PSNR and $0.0741$ in SSIM for \textquotedblleft Mixture 3\textquotedblright. The  maximum gains against MCA\ with K-SVD\ trained dictionaries are $1.41$dB and $0.0953$ for \textquotedblleft Mixture 3\textquotedblright\ again. While we realize that PSNR\ and SSIM are not necessarily the right image quality metrics in this scenario, they do demonstrate objectively the improvements that our method brings over the state of the art.

\begin{table*}[t]
\caption{Objective Quality Assessment of the  X-Ray Separation Performance
of Different Methods on Simulated Mixtures.} 
\label{tab:SimulMixtureTable}
\centering 
\tabcolsep=0.05cm
\footnotesize
\begin{tabular}{c|c| cc| cc| cc| cc|cc} 
\hline\hline
 & & \multicolumn{2}{c}{Mixture 1} & \multicolumn{2}{c}{Mixture 2} & \multicolumn{2}{c}{Mixture
3} & \multicolumn{2}{c}{Mixture 4}& \multicolumn{2}{c}{Mixture 5} \\ [0.1ex]\hline
& Image & PSNR [dB] & SSIM & PSNR [dB] & SSIM  & PSNR [dB] & SSIM & PSNR
[dB] & SSIM & PSNR
[dB] & SSIM\\[0.1ex]
\hline 
\multirow{2}{*}{MCA fixed} & X-ray 1 & 25.69 & 0.7941 & 30.87 & 0.9003 &
27.28 & 0.7915 & 27.99 & 0.7972& 26.96 & 0.8473\\ 
  &  X-ray 2 & 25.50 & 0.8134 & 30.73 & 0.8818 & 27.15 & 0.8198 & 27.86 &
0.8628& 26.78 & 0.8068\\
\hline
\multirow{2}{*}{MCA trained} & X-ray 1 & 26.04 & 0.8245 & 31.07 & 0.8381
& 28.13 & 0.7703 & 27.56 & 0.7783& 27.24 & 0.8258\\
  & X-ray 2 & 25.83 & 0.8485 & 31.15 & 0.8189 & 27.23 & 0.6966 & 27.41 &
0.8464& 27.05 & 0.7927\\
\hline
\multirow{2}{*}{Proposed} & X-ray 1 & \textbf{26.21} & \textbf{0.8583} &
\textbf{31.91} & \textbf{0.9072} & \textbf{28.54} & \textbf{0.8656} & \textbf{28.31}
&
\textbf{0.8266}& \textbf{27.34} & \textbf{0.8592}\\
 & X-ray 2 & \textbf{26.00} & \textbf{0.8759} & \textbf{31.75} & \textbf{0.8892}
& \textbf{28.36} & \textbf{0.8859} & \textbf{28.16} & \textbf{0.8921} & \textbf{27.14}
& \textbf{0.8329}\\
\hline\hline 
\end{tabular}
\end{table*}
    
\section{Conclusion}
\label{sec:conclusion}
We have proposed a novel sparsity-based regularization method for source separation guided by side information. Our method learns dictionaries, coupling  registered acquisitions from diverse modalities, and comes both in a single- and multi-scale framework. The proposed method is applied in the separation  of X-ray images of paintings on wooden panels that are painted on both sides, using the photographs of each side as side information. Experiments on real data, consisting of digital acquisitions of the \textit{Ghent Altarpiece} (1432), verify that the use of side information can be highly beneficial in this application. Furthermore, due to the high resolution of the data relative to the restricted patch size, the multi-scale version of the proposed algorithm improves the  quality of the results significantly. We also observed experimentally that omitting the high frequency crack pixels in the dictionary learning process results in smoother and visually more pleasant separation results. Finally, the superiority of our method, compared to the state-of-the-art MCA technique~\cite{bobin2007sparsity,zibulevsky2001blind,abolghasemi2012blind}, was validated visually using real data  and objectively  using simulated X-ray image mixtures.            

\section*{Acknowledgment}           
Miguel Rodrigues acknowledges valuable feedback from Jonathon Chambers.
                                
\bibliographystyle{IEEEtran}
\bibliography{bib_paper}


%

\end{document}